\documentclass[lettersize,journal]{IEEEtran}
\usepackage{amsmath,amsfonts}
\usepackage{algorithmic}
\usepackage{array}
\usepackage[caption=false,font=normalsize,labelfont=sf,textfont=sf]{subfig}
\usepackage{textcomp}
\usepackage{stfloats}
\usepackage{url}
\usepackage{verbatim}
\usepackage{graphicx}
\def\BibTeX{{\rm B\kern-.05em{\sc i\kern-.025em b}\kern-.08em T\kern-.1667em\lower.7ex\hbox{E}\kern-.125emX}}
\usepackage{balance}
\usepackage{lineno,hyperref}
\usepackage{amssymb}
\usepackage{bm}
\usepackage{color}
\usepackage{threeparttable}
\usepackage{cite}
\usepackage{booktabs}
\usepackage{pifont}
\usepackage{array}
\usepackage{multirow}
\usepackage{tabularx}
\usepackage{caption}
\usepackage{algorithm}
\usepackage{placeins}
\usepackage{tabularray}
\usepackage{makecell}
\usepackage{xcolor} 
\makeatletter
\newcommand{\removelatexerror}{\let\@latex@error\@gobble}
\makeatother

\begin{document}
	
	\title{CroBIM-V: Memory-Quality Controlled Remote Sensing Referring Video Object Segmentation}
	
	\author{
		Haochen~Jiang,
		Yuzhe~Sun,
		Zhe~Dong,
		Tianzhu~Liu,~\IEEEmembership{Member,~IEEE},
		and~Yanfeng~Gu,~\IEEEmembership{Senior Member,~IEEE}%
		\thanks{Manuscript received XX xx, 2025; revised XX xx, 2025; accepted XX xx, 2025.}%
		\thanks{This work was supported by the National Natural Science Foundation of China (Special Program) under Grant No. 624B2051.}%
		\thanks{H. Jiang, Y. Sun, Z. Dong, T. Liu and Y. Gu are with the School of Electronics and Information Engineering, Harbin Institute of Technology, Harbin 150001, China. (email: guyf@hit.edu.cn).}%
	}
	
	\maketitle

\begin{abstract}

Remote sensing video referring object segmentation (RS-RVOS) is challenged by weak target saliency and severe visual information truncation in dynamic scenes, making it extremely difficult to maintain discriminative target representations during segmentation. Moreover, progress in this field is hindered by the absence of large-scale dedicated benchmarks, while existing models are often affected by biased initial memory construction that impairs accurate instance localization in complex scenarios, as well as indiscriminate memory accumulation that encodes noise from occlusions or misclassifications, leading to persistent error propagation.
This paper advances RS-RVOS research through dual contributions in data and methodology. First, we construct RS-RVOS Bench, the first large-scale benchmark comprising 111 video sequences, about 25,000 frames, and 213,000 temporal referring annotations. Unlike common RVOS benchmarks where many expressions are written with access to the full video context, our dataset adopts a strict causality-aware annotation strategy in which linguistic references are generated solely from the target state in the initial frame. Second, we propose a memory-quality-aware online referring segmentation framework, termed Memory Quality Control with Segment Anything Model (MQC-SAM). MQC-SAM introduces a temporal motion consistency module for initial memory calibration, leveraging short-term motion trajectory priors to correct structural deviations and establish accurate memory anchoring. Furthermore, it incorporates a decoupled attention-based memory integration mechanism with dynamic quality assessment, selectively updating high-confidence semantic features while filtering unreliable information, thereby effectively preventing error accumulation and propagation. Extensive experiments on RS-RVOS Bench demonstrate that MQC-SAM achieves state-of-the-art performance.

\end{abstract}

\begin{IEEEkeywords}
	Referring video object segmentation, remote sensing, benchmark dataset, cross-modal.
\end{IEEEkeywords}

\section{Introduction}

\begin{figure}[tbp]
	\begin{center}
		\centerline{\includegraphics[width=1\linewidth]{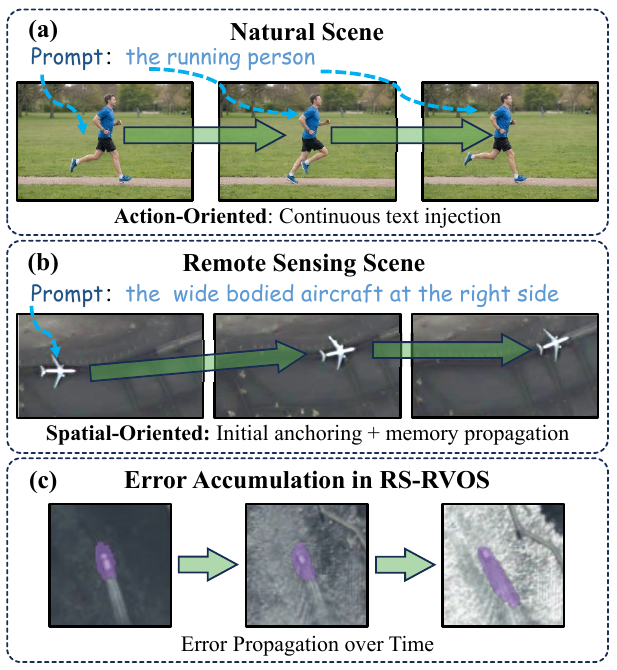}}
		\caption{Key differences between natural-scene RVOS and RS-RVOS. (a) Natural videos benefit from strong temporal action cues. (b) Remote-sensing RVOS relies on initial anchoring and memory propagation under online causality. (c) Errors can accumulate over time, causing drift.}\label{introduction}
	\end{center}
\end{figure}

\IEEEPARstart{R}{}emote sensing video has become increasingly important in applications such as earth observation, disaster monitoring, and traffic management~\cite{kaku2019disaster,zhao2023vehicle,yin2021viso}. Referring video object segmentation (RVOS), which bridges natural language and visual understanding, holds significant potential for intelligent interpretation of remote sensing imagery~\cite{seo2020urvos,khoreva2018refexp,wu2022referformer,botach2022mttr}. However, due to limitations in computational resources and communication bandwidth, remote sensing video analysis is increasingly adopting an online processing paradigm~\cite{hu2023edge}. Under this setting, segmentation decisions must be made frame by frame during observation, rather than after the complete video sequence has been acquired. As summarized in Fig.~\ref{introduction}, this online constraint fundamentally reshapes how text prompts are formulated, how targets are tracked, and how errors accumulate over time in remote sensing RVOS~\cite{wu2023onlinerefer}.

The online paradigm imposes strict causality constraints: at any time, the system can access only the current frame and past information, without recourse to future-frame context. This constraint directly governs the formulation of text prompts. In online scenarios, text descriptions can only be generated at the initial observation stage and must remain fixed throughout subsequent inference~\cite{wu2023onlinerefer,hu2023edge}. In contrast, existing RVOS benchmarks are largely constructed on natural-scene videos, where many expressions are written with access to the full video context, and only a subset provides first-frame expressions~\cite{seo2020urvos,khoreva2018refexp}. However, in the remote sensing domain, a dedicated RVOS benchmark with first-frame-only referring annotations has not yet been systematically established, which limits the development of RS-RVOS methods.

A closer examination reveals a key difference in what can be reliably inferred from the initial frame in natural scenes versus remote sensing scenarios. In many natural-scene videos, the target’s action or state is often observable or strongly suggested even from the initial frame, allowing referring expressions to include dynamic cues for identity specification~\cite{ding2025mevis}. In remote sensing scenarios, however, this assumption often does not hold. Owing to top-down viewpoints, small target sizes, and limited imaging resolution, individual frames in remote sensing videos rarely provide reliable cues about a target’s specific behavior or motion state~\cite{yin2021viso,zhao2023vehicle}. At the initial observation moment, the system cannot predict future trajectories, nor can it reliably infer the target’s ongoing activity from a single frame. In this work, we focus on remote sensing videos where targets exhibit discernible temporal motion patterns, for which motion can serve as an additional verification cue~\cite{yin2021viso}. Consequently, remote sensing prompts should primarily describe spatial location and category attributes for initial identity anchoring, rather than persistent action semantics~\cite{seo2020urvos}.

This paradigm shift has important implications. As illustrated in Fig.~\ref{introduction}(b), directly applying existing RVOS pipelines that repeatedly inject or enforce text semantics across time in online remote sensing scenarios introduces semantic inconsistencies when the target moves away from its initially described location, as location-dependent phrases become invalid and may misguide cross-modal alignment~\cite{wu2022referformer,botach2022mttr}. Meanwhile, even when text is restricted to first-frame identity anchoring, accurately establishing an initial memory anchor remains highly challenging in remote sensing videos. The target often appears small and visually subtle against complex backgrounds~\cite{yin2021viso,zhao2023vehicle}. Such biased or noisy initialization can severely compromise subsequent tracking and segmentation.

As illustrated in Fig.~\ref{introduction}(c), early ambiguities and low-quality predictions cannot be retrospectively corrected using future context. Once such unreliable outputs are written into the memory bank, they contaminate the stored representations and subsequently misguide later-frame inference, resulting in persistent error propagation over time~\cite{oh2019stm,cheng2021stcn,cheng2022xmem}. However, many existing memory-based pipelines update memory with predicted masks/features without reliable quality estimation or explicit mechanisms to reject corrupted updates~\cite{oh2019stm,cheng2022xmem,yang2022deaot}. This motivates the need for robust memory error-correction and quality-control mechanisms in online RS-RVOS.

To address these challenges, we propose a two-stage memory management framework tailored for online remote sensing RVOS. This framework addresses initial memory quality and long-term maintenance through two key insights. First, we identify motion as an orthogonal verification signal; since real moving targets and static distractors exhibit distinct temporal motion patterns, this provides a semantic-independent verification dimension~\cite{yin2021viso}. Accordingly, we design a Temporal Motion-Consistency Initial Memory Calibration module that utilizes adaptive time windows and a motion-semantic dual verification mechanism to refine initial segmentation and suppress noise, providing a high-quality starting point for sequence tracking. Second, we tackle multi-dimensional challenges via decoupled memory management. Given the orthogonal demands of identity consistency, appearance adaptability, and discriminative power, we introduce a Decoupled Attention Memory Integration mechanism~\cite{yang2022deaot,yang2021aot}. This decomposes memory management into three functional dimensions: Cross-modal Semantic Alignment Attention for fixed identity anchoring via fused text and motion priors~\cite{seo2020urvos,wu2022referformer}; Short-term Spatiotemporal Evolution Attention for capturing target dynamics through sliding windows with occlusion-aware gating~\cite{cheng2021stcn,cheng2022xmem}; and Discriminative Error-correction Attention for storing prototypes from high-confusion scenes to prevent false storage~\cite{oh2019stm,yang2022deaot}. These dimensions maintain independent feature manifolds and collaborate via dedicated attention mechanisms to ensure representation reliability across long sequences.

To facilitate research into causality-aware RVOS, we present the RS-RVOS Bench dataset. Diverging from the global annotation paradigm, RS-RVOS Bench employs a frame-specific spatial-semantic generation process, ensuring all prompts are strictly derived from the initial frame to satisfy online causal constraints~\cite{seo2020urvos}. The dataset comprises 111 video sequences with 25,000 frames and 213,000 temporal annotations, with resolutions spanning from 502$\times$512 to 2160$\times$1080. It features 11 fine-grained semantic categories and incorporates diverse challenging scenarios, such as complex interference and dynamic occlusion, screened via a visual discriminative scoring model to provide a systematic evaluation benchmark.

In summary, the contributions of this work can be summarized in the following four aspects:

\begin{itemize}
	\item[(1)] We explicitly differentiate localization-centric remote sensing RVOS from action-centric natural scene RVOS, identifying the semantic drift problem caused by continuous text injection in online settings.
	
	\item[(2)] We propose a two-stage memory management framework that systematically addresses initial quality and long-term maintenance, innovatively using motion as an orthogonal verification signal and decomposing memory into functional dimensions.
	
	\item[(3)] We construct the first systematic benchmark for referring object segmentation in remote sensing videos, RS-RVOS Bench, which significantly advances existing datasets in scale, annotation density, and task formulation, and provides a unified evaluation platform for RS-RVOS research.
	
	\item[(4)] Extensive experiments on RS-RVOS Bench demonstrate that our framework achieves significant performance gains over state-of-the-art methods in challenging remote sensing scenarios, validating its effectiveness and robustness.
\end{itemize}

The remainder of this paper is organized as follows. Section~\ref{section:RelatedWork} reviews related work on referring object segmentation in remote sensing images, remote sensing video object segmentation, and referring video object segmentation. Section~\ref{section:dataset} presents the RS-RVOS Bench dataset, describing its causality-aware annotation paradigm, visual discriminability-based sequence selection, and dataset statistics. Section~\ref{section:methods} details the proposed MQC-SAM framework, including the temporal motion-consistent initial memory calibration module and the decoupled attention-based memory integration mechanism.  Section~\ref{section:EXPERIMENTS} provides a comprehensive evaluation of MQC-SAM, with comparisons against state-of-the-art methods and ablation studies on key components. Finally, Section~\ref{sec:conclusion} concludes the paper and outlines future research directions.

\section{Related Work}
\label{section:RelatedWork}

\subsection{Referring Segmentation in Remote Sensing Images}

Referring image segmentation has made substantial progress in natural-image domains, where existing methods leverage vision--language alignment to localize and segment target instances guided by textual descriptions.
Early representative works, such as LAVT~\cite{yang2022lavt}, established Transformer-based cross-modal interaction frameworks to bridge linguistic expressions and visual representations.
Subsequent approaches, including CRIS and DiffRIS~\cite{wang2022cris,dong2025diffris}, further improved text-to-pixel alignment by exploiting large-scale vision--language pretraining (e.g., CLIP~\cite{radford2021clip}).
In parallel, referring expression comprehension (REC) also benefited from contrastive learning and multimodal pretraining (e.g., UNITER~\cite{chen2020uniter}, ALBEF~\cite{li2021albef}), offering insights into cross-modal grounding beyond segmentation.

Extending referring segmentation to remote sensing imagery, however, required addressing domain-specific challenges.
Unlike natural images, targets in remote sensing scenes were often sparsely distributed and occupied only a small fraction of the large-area coverage, which could weaken conventional attention mechanisms and aggravate background dominance.
Recent studies began to explore this direction.
RMSIN~\cite{liu2024rmsin} targeted small-object and weak-saliency issues by introducing rotated multi-scale interaction to enhance fine-grained representations, and released the RRSIS-D benchmark for systematic evaluation~\cite{liu2024rmsin}.
CroBIM~\cite{dong2024crobim} mitigated background interference by explicitly modeling object--background interactions and further contributed the large-scale RISBench benchmark for RRSIS.
CroBIM-U~\cite{sun2026crobim} introduced a pixel-wise referring-uncertainty map to adaptively gate language fusion and local refinement, improving robustness and boundary precision in referring remote-sensing image segmentation.
In addition, prompt-based adaptation of segmentation foundation models was explored for remote sensing instance segmentation; for example, RSPrompter~\cite{chen2024rsprompter} learned to generate prompts that better exploit the Segment Anything Model (SAM)~\cite{kirillov2023sam} under remote sensing conditions.

Beyond method development, remote sensing vision--language benchmarks also emerged to facilitate broader evaluation across tasks and scenarios, covering applications such as urban monitoring and disaster assessment (e.g., captioning, grounding, and VQA)~\cite{lu2017rsicd,zhan2023rsvg,lobry2020rsvqa,li2024vrsbench}.

Despite these advances, most existing remote sensing referring segmentation methods remain limited to static images and are typically annotated with access to complete spatial context and global scene semantics, implicitly assuming full-image observability.
This setting deviates from practical remote sensing deployments, where causality constraints often require descriptions to be formulated solely from the initial observable frame.
Moreover, weak target saliency makes reliable single-frame initialization challenging; in video sequences, inaccurate initialization can further propagate errors and impair subsequent tracking over time.

\subsection{Remote Sensing Video Object Segmentation}

Remote sensing video object segmentation is fundamental for dynamic Earth observation applications such as traffic monitoring, disaster tracking, and maritime surveillance~\cite{li2023satmtb,zhao2022satsot,kou2024sat}.
Conventional approaches exploited motion cues and temporal consistency to preserve object identities across frames.
Early techniques, including optical-flow-based tracking and Kalman filtering, provided foundational solutions but often struggled under complex backgrounds and intermittent visibility.

Deep learning substantially advanced this field through memory-based architectures.
STM~\cite{oh2019stm} introduced a paradigm that maintained a feature memory to encode historical information, enabling effective retrieval of relevant patterns in subsequent frames.
Building upon this framework, STCN~\cite{cheng2021stcn} improved space--time correspondence modeling with more effective memory matching, while AOT~\cite{yang2021aot} enhanced multi-object association through transformer-based propagation.
XMem~\cite{cheng2022xmem} further proposed a long-term memory design with multiple memory stores at different temporal scales, improving robustness over long video sequences.

Remote sensing video segmentation presented challenges that differed from those in natural-video analysis.
Observation gaps caused by cloud~\cite{jin2025vcdformer} cover or imaging limitations could lead to intermittent target visibility, while complex terrain textures and man-made structures introduced persistent ambiguity~\cite{li2023satmtb}.
Accordingly, several studies explored domain-specific adaptations to alleviate these effects.

Nevertheless, most existing remote sensing video segmentation methods assumed predefined target identities and required manual initialization with bounding boxes or masks in the first frame, limiting their applicability in fully autonomous monitoring scenarios.
Moreover, purely visual tracking systems lacked explicit semantic grounding and could be susceptible to identity drift under significant appearance variations.
Incorporating language guidance remained relatively underexplored in remote sensing videos, yet it holds promise for improving automation and robustness through explicit semantic anchoring.
\begin{figure*}[tbp]
	\begin{center}
		\centerline{\includegraphics[width=1\linewidth]{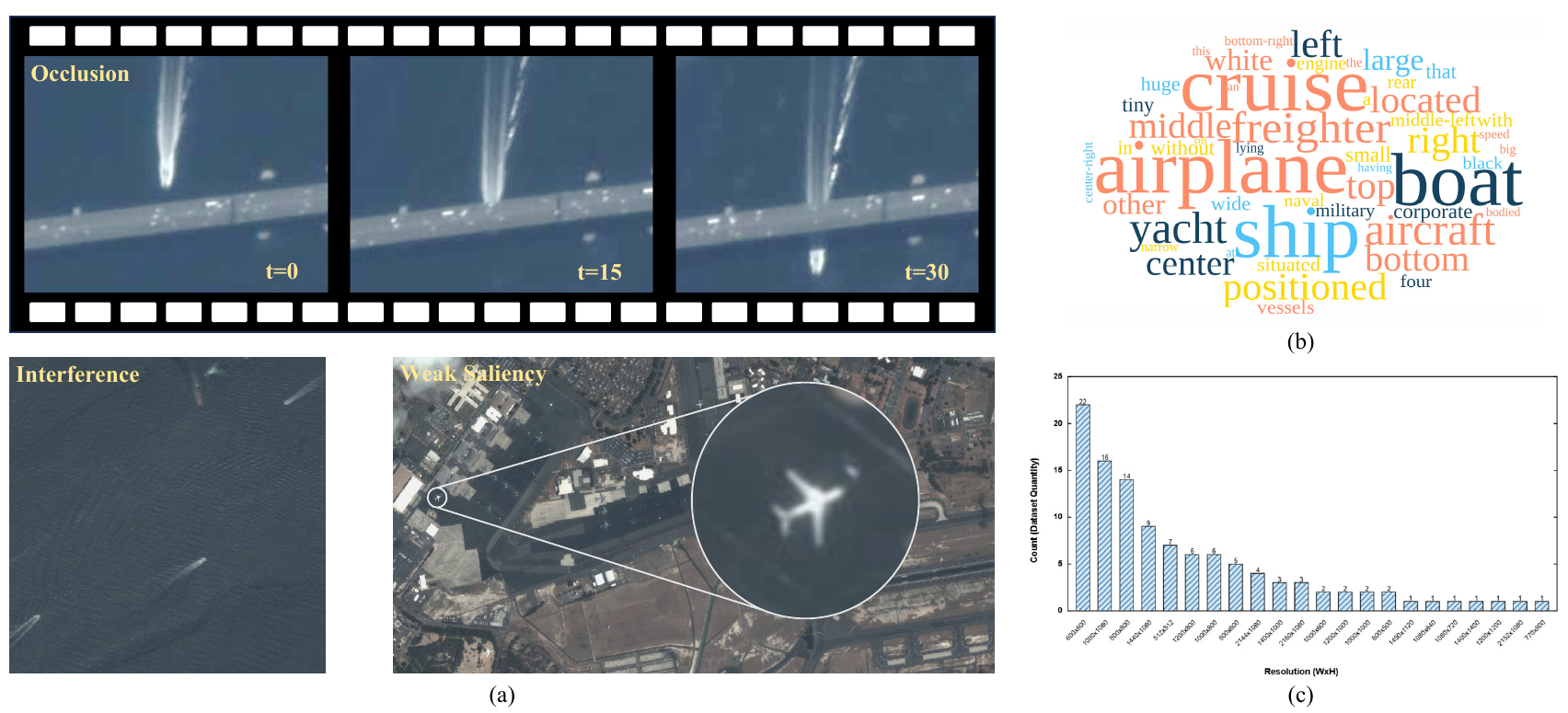}}
		\caption{Representative characteristics of the RS-RVOS Bench dataset. 
			(a) Visual examples illustrating intrinsic challenges in remote sensing videos, including occlusion over time, background interference from visually similar distractors, and weak target saliency under large-scale scenes. 
			(b) Word cloud of referring expressions, showing the distribution of spatial terms and attribute descriptions used for causality-compliant prompt generation. 
			(c) Statistical distribution of video resolutions in RS-RVOS Bench.}
		\label{dataset2}
	\end{center}
\end{figure*}

\begin{figure}[t]
	\centering
	\includegraphics[width=\linewidth]{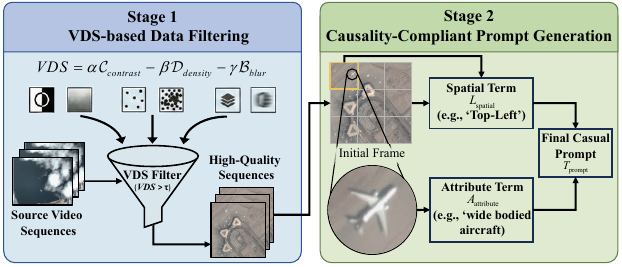}
	\caption{Overview of the RS-RVOS Bench dataset construction pipeline. 
		Stage~1 applies VDS-based filtering to select visually discriminative sequences.
		Stage~2 generates causality-compliant referring expressions from the initial frame.}
	\label{fig:dataset1}
\end{figure}

\subsection{Referring Object Segmentation in Remote Sensing Videos}

Referring video object segmentation (RVOS) integrates language understanding with video analysis and has become a prominent paradigm in natural-scene understanding.
Early representative works, such as ReferFormer~\cite{wu2022referformer} and MTTR~\cite{botach2022mttr}, demonstrated end-to-end architectures that jointly modeled textual queries and temporal video features through multimodal Transformers for accurate localization and segmentation.
Subsequent studies further improved temporal grounding and cross-modal alignment, for example via video-level alignment enhancement (e.g., SOC~\cite{luo2023soc}), online query propagation (e.g., OnlineRefer~\cite{wu2023onlinerefer}), or by leveraging stronger vision--language foundations and segmentation backbones.

Benchmarks in natural scenes also reflected different annotation assumptions.
For instance, Refer-YouTube-VOS provided both full-video expressions and first-frame expressions~\cite{seo2020urvos}, while newer datasets such as MeViS emphasized motion expressions to strengthen temporal reasoning~\cite{ding2023mevis}.
These settings were generally compatible with natural-scene videos, where action/state cues were often observable or inferable from early observations.

In contrast, referring object segmentation in remote sensing videos remains at an early stage.
To the best of our knowledge, a dedicated RS-RVOS benchmark with referring annotations for remote sensing videos has not yet been established.
Existing remote sensing referring datasets remained predominantly image-based, leaving the video-level setting underexplored.
Inherent challenges of remote sensing imagery---including weak target saliency, severe observation gaps, and complex backgrounds---called for methodological frameworks beyond those developed for natural scenes.
Existing studies predominantly relied on natural-scene benchmarks, whose task formulations and model designs were poorly aligned with the requirements of online remote sensing applications.
As a result, RVOS paradigms tailored to natural scenes could exhibit intrinsic incompatibilities when applied to remote sensing scenarios.

\section{Dataset Construction and Analysis}
\label{section:dataset}

\subsection{RS-RVOS Bench Dataset Construction}
\label{subsection:dataset_construction}

At present, the remote sensing community lacks a publicly available benchmark specifically designed for the task of referring video object segmentation in satellite videos. Existing referring video segmentation and video object segmentation benchmarks are predominantly derived from natural scenes (e.g., DAVIS~\cite{perazzi2016davis}, YouTube-VOS~\cite{xu2018youtubevos}, and RVOS benchmarks such as URVOS~\cite{seo2020urvos} and MeViS~\cite{ding2023mevis}), whose task formulations and data organization strategies fail to adequately reflect the characteristics of remote sensing video analysis. To fill this gap, we construct RS-RVOS Bench, a dedicated dataset tailored for referring video object segmentation in remote sensing scenarios.

RS-RVOS Bench is built upon the SAT-MTB satellite video repository~\cite{li2023satmtb}. To ensure suitability for instance-level segmentation and maintain high data quality, we first curate the raw video sequences by removing scenes suffering from severe motion blur or excessively high target density, which often leads to insufficient visual discriminability.
To quantitatively assess candidate sequences, we introduce a Visual Discriminability Score (VDS), defined as:
\begin{equation}
	VDS(V) = \lambda_c \cdot \mathcal{C}_{\text{contrast}}(V)
	- \lambda_d \cdot \mathcal{D}_{\text{density}}(V)
	- \lambda_b \cdot \mathcal{B}_{\text{blur}}(V),
\end{equation}
where $\mathcal{C}_{\text{contrast}}(V)$ measures target--background separability on the first annotated frame by the mean feature (or intensity) difference between the target region and a narrow surrounding band,
$\mathcal{D}_{\text{density}}(V)$ is the average number of annotated instances per frame,
and $\mathcal{B}_{\text{blur}}(V)$ measures the overall blur level of a sequence.
Specifically, we first compute a sharpness score using the variance of the Laplacian~\cite{pech2000diatom,pertuz2013focus}:
\begin{equation}
	\mathcal{S}_{\text{lap}}(V)=\frac{1}{T}\sum_{t=1}^{T}\mathrm{Var}\!\left(\nabla^2 I_t\right),
\end{equation}
and normalize it to $[0,1]$ over all candidate videos, denoted as $\widehat{\mathcal{S}}_{\text{lap}}(V)$.
We then convert it into a blur score by:
\begin{equation}
	\mathcal{B}_{\text{blur}}(V)=1-\widehat{\mathcal{S}}_{\text{lap}}(V)
\end{equation}

To make the three terms comparable, we normalize $\mathcal{C}{\text{contrast}}(V)$, $\mathcal{D}{\text{density}}(V)$, and $\mathcal{S}_{\text{lap}}(V)$ to $[0,1]$ over all candidate videos before aggregation.

For textual annotation, RS-RVOS Bench adopts a frame-specific spatial--semantic generation paradigm to accommodate the characteristics of remote sensing video analysis.
Unlike global description strategies that rely on complete video sequences, all referring expressions in RS-RVOS Bench are generated strictly based on the visible state of the initial frame $I_0$ to satisfy the online causal constraint~\cite{seo2020urvos}.
Formally, the text generation function $\mathcal{G}_{\text{text}}$ is defined as:
\begin{equation}
	T_{\text{prompt}} = \mathcal{G}_{\text{text}}(L_{\text{spatial}}, A_{\text{attribute}} \mid I_0),
\end{equation}
where $L_{\text{spatial}}$ denotes the spatial description component and $A_{\text{attribute}}$ denotes the attribute description component.
Spatial descriptions are generated by partitioning the initial frame into a $3 \times 3$ grid and assigning explicit locational terms according to the target centroid $(x_c, y_c)$.
Attribute descriptions are constructed from fine-grained semantic categories of the target, providing a stable and consistent referring anchor without introducing information from future frames.

\subsection{Dataset Statistics and Characteristics}
\label{subsection:dataset_statistics}

RS-RVOS Bench consists of 111 satellite video sequences, covering approximately 25,000 frames in total.
The dataset provides 213,000 temporal referring annotations, each associated with a precise instance-level segmentation mask.
The dataset is split at the sequence level into a training set containing 82 videos and a test set containing 29 videos.

The dataset preserves the intrinsic diversity of satellite sensor observations, with video resolutions ranging from $502 \times 512$ to $2,160 \times 1,080$ pixels.
Semantic labels are finely categorized into 11 distinct classes, enabling models to learn and discriminate subtle semantic differences among targets.
Moreover, RS-RVOS Bench deliberately retains a wide range of challenging real-world scenarios, including complex backgrounds with multi-source occlusions and scenes with weak target saliency, which are commonly reported as intrinsic difficulties in satellite video understanding benchmarks~\cite{li2023satmtb}.
These characteristics collectively establish RS-RVOS Bench as a challenging and representative benchmark for evaluating referring video object segmentation methods in remote sensing applications.

\begin{figure*}[tbp]
	\begin{center}
		\centerline{\includegraphics[width=1\linewidth]{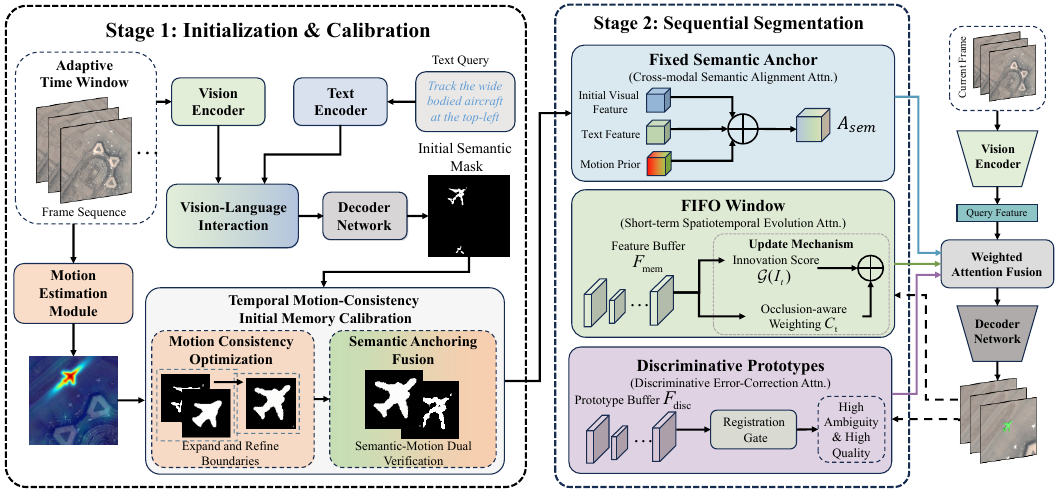}}
		\caption{Overview of the proposed MQC-SAM framework for online remote sensing referring video object segmentation. The framework follows a two-stage design. The left part illustrates Stage~1: Initialization and Calibration, where vision--language interaction and temporal motion-consistency calibration are jointly employed to generate a high-quality initial memory, leveraging adaptive time windows and motion--semantic dual verification to refine initial segmentation and suppress noise. The right part depicts Stage~2: Sequential Segmentation, which performs online inference through a Decoupled Attention Memory Integration mechanism, including a fixed cross-modal semantic anchor for identity consistency, a short-term spatiotemporal FIFO window for appearance adaptation, and discriminative prototypes for error correction. These memory components are fused via weighted attention to enable robust target tracking under strict online causality constraints.}
		\label{flowchart}
	\end{center}
\end{figure*}

\section{Methodology}
\label{section:methods}

\subsection{Method Overview}

Given a sequence of consecutive remote sensing video frames
$\{I_t\}_{t=0}^{T}$ acquired over the same geographic region and a textual
query $Q$ describing the target category or attributes in the initial frame,
the objective of referring video object segmentation is to generate a sequence
of frame-wise segmentation masks $\{R_t\}_{t=0}^{T}$ that consistently delineate
the referred target throughout the sequence~\cite{seo2020urvos,wu2022referformer,botach2022mttr}.

Unlike offline paradigms commonly adopted in natural-scene settings, we follow
an online formulation tailored for remote sensing scenarios. The textual query
$Q$ is provided only at the initial frame $I_0$ to establish a semantic anchor,
while segmentation in subsequent frames relies entirely on memory representations
initialized from $I_0$ and propagated over time~\cite{wu2023onlinerefer,oh2019stm,cheng2022xmem}.

Remote sensing imagery exhibits characteristics that fundamentally challenge
conventional segmentation methods~\cite{li2023satmtb,yin2021viso}. Targets typically occupy only a tiny fraction
of wide-area imagery and lack visually salient appearances. Let $A_{\text{target}}$
denote the area of the target region and $A_{\text{image}}$ the total image area.
Target saliency can be defined as:
\begin{equation}
	\mathcal{S}_{\text{sal}} = \frac{A_{\text{target}}}{A_{\text{image}}} \ll 1
\end{equation}

This weak saliency is further exacerbated by complex background clutter.
Moreover, occlusions caused by clouds, bridges, and other structures lead to
intermittent visibility disruptions, often breaking target correspondence in
traditional tracking-based pipelines~\cite{li2023satmtb}.

To address these challenges, robust memory mechanisms are required to accumulate
and preserve reliable target representations over time~\cite{oh2019stm,cheng2021stcn,cheng2022xmem,yang2022deaot}. Let $\mathcal{M}$ denote
the memory bank, whose state at time $t$ is given by:
\begin{equation}
	\mathcal{M}_t = \{\mathbf{m}_i\}_{i=1}^{N_t},
\end{equation}
where $\mathbf{m}_i$ represents an individual memory feature. When properly
maintained, the memory bank serves as a stable reference that anchors the
segmentation process even when instantaneous visual evidence is ambiguous or
incomplete~\cite{oh2019stm,cheng2022xmem}. However, effective memory design must address two critical issues:
the quality of initial memory representations and the prevention of error
accumulation during temporal propagation~\cite{cheng2022xmem,yang2022deaot}.

To this end, we propose MQC-SAM, a two-stage memory management framework. In the initialization
stage, the initial frame $I_0$ and the textual query $Q$ are encoded by visual
and textual encoders, respectively, and fused to produce an initial segmentation
mask $R_0$. This preliminary result is then rigorously validated and refined
through a temporal motion-consistency-based memory calibration module.

During sequential segmentation, video frames are processed in temporal order,
while a decoupled attention-based memory integration mechanism governs memory
updates. This mechanism decomposes memory management into three functionally
orthogonal attention dimensions, each maintaining an independent feature
manifold to address different aspects of the segmentation problem~\cite{vaswani2017transformer,yang2022deaot}. By enforcing
strict memory registration criteria, only predictions that satisfy reliability
thresholds are allowed to update the memory bank, effectively preventing
progressive drift caused by error accumulation~\cite{cheng2022xmem,yang2022deaot}. Together, these two stages ensure
that segmentation starts from a reliable initialization and remains stable
throughout the entire sequence.

\subsection{Temporal Motion Consistency for Initial Memory Calibration}

In RVOS for remote sensing, the quality of initial segmentation critically determines performance across the entire sequence~\cite{oh2019stm,cheng2022xmem}. Due to weak object saliency in remote sensing scenes, initial masks often suffer from incomplete structures or contamination by texturally similar backgrounds~\cite{li2023satmtb,yin2021viso}. Once encoded into memory, these defects persistently propagate errors throughout subsequent frames. The proposed Temporal Motion Consistency Calibration module (TMCC) addresses this issue through a key insight: genuine moving objects exhibit fundamentally different temporal motion patterns compared to static backgrounds or distractors in remote sensing videos. This motion information provides an independent verification signal orthogonal to semantic segmentation, effectively compensating for the limitations of single-frame semantic analysis in remote sensing scenarios. By integrating temporal motion cues with initial semantic predictions, our module achieves both structural refinement and noise suppression in initial memory.

Given that remote sensing objects may remain stationary initially, computing motion between adjacent frames may yield insufficient signal. We therefore employ an adaptive temporal window mechanism to capture adequate motion evidence. Let $R_0$ denote the initial segmentation region. We search for the optimal observation window length $n^*$ that enables detection of significant motion within the target region:
\begin{equation}
	n^* = \arg\min_{n} \left\{ n : \exists \text{ region} \in R_0, \bar{d}(\text{region}, n) > \tau_{\text{motion}} \right\},
\end{equation}
where $\bar{d}(\text{region}, n)$ represents the average displacement magnitude of a region over an $n$-step temporal interval (i.e., using $n$ consecutive inter-frame displacements), and $\tau_{\text{motion}}$ is the motion detection threshold. Starting from initial window $n_0$, the mechanism incrementally extends by $\Delta n$ steps until detecting valid motion or reaching maximum window $n_{\max}$.

Having determined the optimal window, we compute dense inter-frame displacement fields and perform temporal averaging to obtain robust motion estimates. Let $\mathbf{d}_t$ denote the displacement vector field between frames $t$ and $t+1$. We define a temporally averaged motion-intensity map as:
\begin{equation}
	D(x,y) = \frac{1}{n^*} \sum_{t=0}^{n^*-1} \|\mathbf{d}_t(x,y)\|_2,
\end{equation}

This field encodes pixel-wise motion intensity across the observation window, providing pixel-level evidence for subsequent mask calibration.

\begin{figure}[tbp]
	\begin{center}
		\centerline{\includegraphics[width=\linewidth]{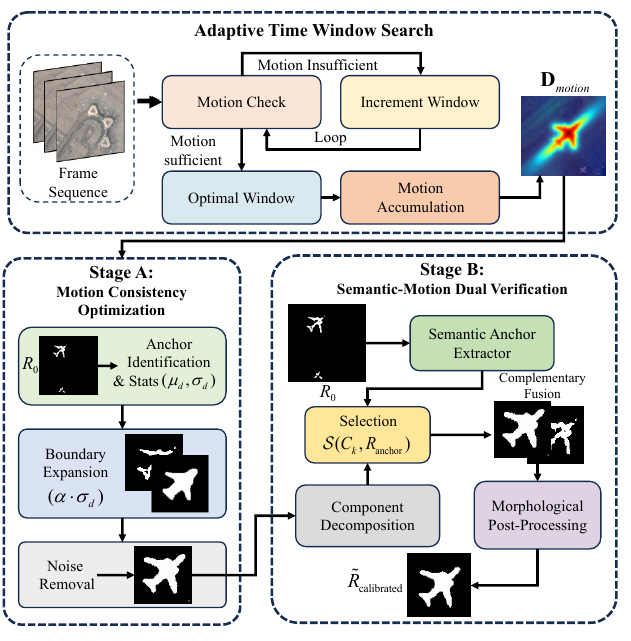}}
		\caption{Temporal motion-consistency initial memory calibration module. 
			The module is followed by motion consistency optimization and semantic--motion dual verification to refine the initial mask and generate the calibrated memory $\tilde{R}_{\text{calibrated}}$.}
		\label{fig:innovation1}
	\end{center}
\end{figure}

\subsubsection{Motion Consistency Optimization}

Using the temporal displacement field, we employ a two-stage calibration process. The first stage adaptively expands or prunes mask boundaries based on target motion statistics. We first identify reliably moving regions $R_{\text{moving}}$ from the initial mask as anchors, computing their motion statistics:
\begin{subequations}
	\begin{align}
\mu_d &= \mathbb{E}[D\mid R_{\text{moving}}] \\
\sigma_d &= \text{Std}[D\mid R_{\text{moving}}]
	\end{align}
\end{subequations}

We then incorporate neighboring pixels with consistent motion patterns to expand the mask:
\begin{equation}
	\begin{aligned}
		R_{\text{expand}}
		&= R_{\text{moving}} \cup
		\Bigl\{ (x,y) \in \mathcal{N}(R_{\text{moving}}) : \\
		&\qquad \bigl|D(x,y) - \mu_d\bigr|
		< \alpha \cdot \sigma_d \Bigr\},
	\end{aligned}
\end{equation}
where $\mathcal{N}(R_{\text{moving}})$ denotes the spatial neighborhood of moving regions, and $\alpha$ controls expansion scope. This recovers under-segmented structures caused by weak saliency. Subsequently, we remove outlier regions with inconsistent motion:
\begin{equation}
	R_{\text{refine}} = R_{\text{expand}} \setminus \left\{ (x,y) : |D(x,y) - \mu_d| > \beta \cdot \sigma_d \right\},
\end{equation}
where $\beta > \alpha$ serves as the outlier removal threshold, eliminating incorrectly included static backgrounds or motion-inconsistent distractors.

\subsubsection{Semantic Anchoring Fusion}

The second stage preserves the class discriminative power of initial semantic predictions while leveraging motion refinement. Although motion consistency analysis effectively identifies coherently moving pixel groups and recovers structural completeness, it operates on low-level physical features without high-level semantic information. Critically, motion-based refinement may erroneously discard object regions lacking motion saliency or introduce dynamic background elements with similar motion magnitudes. A synergistic framework between semantic segmentation and motion analysis is therefore essential for mutual verification and completion~\cite{wu2022referformer,botach2022mttr}. Let $R_{\text{anchor}}$ denote high-confidence semantic regions from $R_0$ exceeding threshold $\tau_{\text{sem}}$, and $R_{\text{refine}}$ the motion-optimized region. We propose a bidirectional verification mechanism that filters dynamic distractors using semantic anchors while recovering static object structures erroneously removed during motion optimization.

Specifically, let $R_{\text{refine}}$ contain $K$ spatially independent connected components $\{C_k\}_{k=1}^{K}$. To select semantically consistent regions while excluding dynamic background interference, we compute spatial correspondence between each candidate and semantic anchors, integrating region overlap and boundary proximity:
\begin{equation}
	\begin{aligned}
		\mathcal{S}(C_k, R_{\text{anchor}})
		&= \frac{|C_k \cap R_{\text{anchor}}|}
		{\min\!\bigl(|C_k|, |R_{\text{anchor}}|\bigr)} \\
		&\quad + \lambda \cdot \exp\!\left(
		-\frac{d_{\partial}(C_k, R_{\text{anchor}})}{\sigma_s}
		\right)
	\end{aligned}
\end{equation}

The first term quantifies spatial overlap intensity, identifying motion regions semantically aligned with the target. The second term captures spatial proximity through boundary distance $d_{\partial}$, retaining motion-completed regions belonging to the object but missed by semantic segmentation. $\lambda$ balances the two terms, and $\sigma_s$ scales distance. We select the most semantically associated candidate as the motion-verified region:
\begin{equation}
	k^* = \arg\max_{k \in \{1,\ldots,K\}} \mathcal{S}(C_k, R_{\text{anchor}}), \quad
	R_{\text{motion}} = C_{k^*}
\end{equation}

This filtering effectively excludes spatially unrelated dynamic distractors, ensuring semantic purity of motion-based completion.

Given semantic anchor $R_{\text{anchor}}$ and motion-verified region $R_{\text{motion}}$, we adopt a complementary fusion strategy to generate the final calibrated mask. This strategy exploits complementary strengths: semantic anchors provide class-specific anchoring to preserve static structures erroneously removed by motion optimization, while motion-verified regions contribute temporally validated structural completeness that recovers under-segmented boundaries missed due to weak saliency. Their spatial integration is formulated as:
\begin{equation}
	R_{\text{calibrated}} = R_{\text{anchor}} \cup R_{\text{motion}}
\end{equation}

This fusion mechanism offers three key advantages: spatially overlapping regions receive mutual confirmation with highest confidence; static structures in semantic anchors erroneously removed by motion optimization are recovered; potential dynamic distractors in motion-verified regions are excluded during filtering due to lack of spatial correspondence with semantic anchors.

To enhance spatial coherence, we apply morphological post-processing. Closing operations bridge internal gaps caused by motion estimation noise, and hole-filling ensures topological integrity. Let $\Phi_{\text{morph}}$ denote the morphological operator. The final calibrated memory is:
\begin{equation}
	\tilde{R}_{\text{calibrated}} = \Phi_{\text{morph}}(R_{\text{calibrated}})
\end{equation}

After temporal motion-consistency calibration, the initial memory possesses three key characteristics: structural integrity is significantly enhanced, completing structures that were missed due to weak target saliency; background noise is effectively suppressed, removing static distractors and falsely segmented regions with similar textures; meanwhile, dual anchoring of semantics and motion is maintained, taking into account both category and orientation discriminability as well as motion consistency. This module outputs a high-quality initial mask $\tilde{R}_{\text{calibrated}}$.
It will be used to construct fixed cross-modal semantic anchors. By deeply fusing the calibrated mask with the visual features of the initial frame $I_0$,
the model can simultaneously encode the target’s visual appearance, spatial position, and structurally complete information verified by motion in the initial state.
This fusion process establishes a “complete semantic representation of the target under the current scene background,” providing a stable multimodal reference baseline for segmentation of the entire sequence and effectively reducing cumulative errors caused by initial memory bias.

\subsection{Decoupled Attention Memory Integration Mechanism}

After initial memory calibration, the primary challenge in sequence segmentation lies in maintaining memory bank reliability during long-term temporal reasoning. Remote sensing video object segmentation presents unique complexities: complex background clutter causes target confusion, occlusion leads to intermittent feature loss, and cumulative errors induce long-range target drift. To address these challenges, we propose a Decoupled Attention Memory Integration mechanism (DAMI) that partitions memory management into three functionally orthogonal attention dimensions~\cite{vaswani2017transformer,yang2022deaot,cheng2022xmem}. Each dimension independently maintains a specific feature manifold with dedicated retrieval. Specifically, cross-modal semantic alignment attention anchors global target identity via fixed semantic references; short-term spatio-temporal evolution attention adapts to local appearance variations through sliding temporal windows; discriminative refinement attention distinguishes distractors by storing high-ambiguity discriminative prototypes. These three dimensions synergistically enhance memory robustness from orthogonal perspectives: semantic stability, temporal adaptability, and distractor discriminability.

During inference at each frame, feature fusion integrates the three attention dimensions. 
Let $\mathbf{f}_t$ denote the visual feature extracted from frame $I_t$, and let $\mathbf{q}_t = \phi_q(\mathbf{f}_t)$ denote the corresponding query embedding used in attention, where $\phi_q(\cdot)$ is a learnable projection. 
We use $\mathbf{m}_i$ to denote an entry stored in the memory bank.
The fused representation is computed as:
\begin{equation}
	\begin{aligned}
		\mathbf{F}_{\text{fused}}
		&= \omega_1 \cdot \text{Attn}_{\text{sem}}(\mathbf{q}_t, \mathbf{A}_{\text{sem}}) \\
		&\quad + \omega_2 \cdot \text{Attn}_{\text{short}}(\mathbf{q}_t, \mathcal{F}_{\text{mem}}) \\
		&\quad + \omega_3 \cdot \text{Attn}_{\text{disc}}(\mathbf{q}_t, \mathcal{F}_{\text{disc}}),
	\end{aligned}
\end{equation}
where $\omega_1, \omega_2, \omega_3$ are learnable fusion weights for cross-modal semantic alignment, short-term spatio-temporal evolution, and discriminative refinement attention, respectively.

Cross-Modal Semantic Alignment Attention. Cross-modal semantic alignment attention constructs a fixed semantic anchor $\mathbf{A}_{\text{sem}}$, providing a stable identity reference for sequence-wide tracking~\cite{wu2022referformer,botach2022mttr}. This anchor is generated at initialization by deeply fusing the initial frame $I_0$ with the temporally motion-consistency--calibrated mask $\tilde{R}_{\text{calibrated}}$:

\begin{equation}
	\mathbf{A}_{\text{sem}} = \text{MemoryEncoder}\left( I_0, \tilde{R}_{\text{calibrated}} \right)
\end{equation}

Through this fusion, the semantic anchor encodes complementary priors from text, vision, and motion. The textual semantic prior arises from the category constraints and spatial orientation information introduced by the text query $Q$ during the generation of $\tilde{R}_{\text{calibrated}}$. The visual appearance prior is provided by the initial frame $I_0$, capturing the target’s texture, shape, and other visual characteristics in the initial scene. The motion-consistency prior is embedded in the spatial distribution of the calibrated mask $\tilde{R}_{\text{calibrated}}$, which is validated through temporal motion verification and reflects structural integrity jointly confirmed by motion and semantics. This multimodal fusion mechanism endows the semantic anchor with semantic interpretability, visual discriminability, and adaptability to dynamic scenes.  
The fixed anchor remains unchanged throughout the entire sequence and serves as a spatiotemporally invariant identity reference. When the target undergoes appearance variation, occlusion, or complex motion, features from subsequent frames can query this anchor to obtain stable identity confirmation, thereby effectively suppressing identity drift caused by single-modality information loss and ensuring robust target locking even under challenging conditions such as weak saliency or dynamic background interference.

Short-Term Spatio-Temporal Evolution Attention. This dimension models non-rigid deformations and radiometric variations in remote sensing videos, ensuring feature continuity and appearance adaptation within local temporal ranges during sequential segmentation. It maintains a first-in-first-out short-term memory buffer $\mathcal{F}_{\text{mem}} = \{\mathbf{f}_{t-L+1}, \ldots, \mathbf{f}_{t-1}\}$ with window-length parameter $L$, forming a sliding buffer over recent frames. To incorporate temporal recency, features are weighted by relative temporal distance from the current frame. Let $\mathbf{P}_{\text{rel}} \in \mathbb{R}^{1 \times (L-1)}$ denote the relative position bias matrix, where the $i$-th element encodes the temporal weight for the $i$-th buffered frame. Attention is computed as:
\begin{equation}
	\text{Attn}_{\text{short}}(\mathbf{q}_t, \mathcal{F}_{\text{mem}}) = \text{Softmax}\left( \frac{\mathbf{q}_t \cdot \mathcal{F}_{\text{mem}}^T}{\sqrt{d}} + \mathbf{P}_{\text{rel}} \right) \cdot \mathcal{F}_{\text{mem}}
\end{equation}

This bias prioritizes recent texture and morphological changes based on temporal proximity.

Feature registration employs dual gating for efficient and reliable updates. The first gate evaluates temporal novelty via an innovation score, quantifying feature changes relative to the short-term memory buffer:
\begin{equation}
\hat{\mathbf{f}}_t = \sum_{k=t-L+1}^{t-1} w_k \mathbf{f}_k, \quad
w_k = \frac{\exp(-\eta (t-k))}{\sum_{\tau=t-L+1}^{t-1} \exp(-\eta (t-\tau))},
\end{equation}
\begin{equation}
	\mathcal{G}(I_t) = \frac{1}{|\Omega|}\sum_{p\in\Omega} \left\| \mathbf{f}_t(p) - \hat{\mathbf{f}}_t(p) \right\|_2
\end{equation}
Here $\Omega$ denotes the set of spatial locations on the feature map. This score measures feature novelty as temporal innovation with respect to a short-term predictor, and $\tau_{\text{gain}}$ is set by a fixed percentile on $\mathcal{G}(I_t)$ over a sequence to avoid manual tuning. Features are registered only when exhibiting significant novelty: $\text{Register}(t) = \mathbb{I}\left[ \mathcal{G}(I_t) > \tau_{\text{gain}} \right]$, filtering redundant frames to improve storage efficiency. 

The second gate applies occlusion-aware weighting using the decoder's confidence map $\mathbf{C}_t \in [0,1]^{H \times W}$ as soft attention:
\begin{equation}
	\mathbf{f}_t^{\text{weighted}} = \mathbf{f}_t \odot \mathbf{C}_t
\end{equation}

This suppresses feature aggregation from low-confidence occluded regions, preventing noise pollution in memory.

Discriminative Refinement Attention. This dimension stores high-ambiguity discriminative prototypes to enhance instance discrimination under strong background interference. It maintains a prototype buffer $\mathcal{F}_{\text{disc}}$ containing temporally unconstrained discriminative priors---whenever similar distractors appear in the field of view, these prototypes activate to aid discrimination regardless of their temporal origin. Unlike short-term attention emphasizing temporal continuity, this pool prioritizes discriminative value.

Feature registration uses dual verification. First, adaptive ambiguity awareness computes distribution divergence between main and alternative hypotheses via KL divergence~\cite{kullback1951kl}:
\begin{equation}
	D_{\text{KL}}(P_{\text{main}} \,||\, P_{\text{alt}}) = \sum_i P_{\text{main}}(i) \log \frac{P_{\text{main}}(i)}{P_{\text{alt}}(i)}
\end{equation}

Low divergence indicates high semantic ambiguity, detecting potential distractors: $\text{Ambiguous}(t) = \mathbb{I}\left[ D_{\text{KL}} < \tau_{\text{amb}}(t) \right]$. Second, representation quality assessment employs a learned robustness scorer $\mathcal{S}_{\text{robust}}(\cdot)$ evaluating spatio-temporal consistency:
\begin{equation}
	s_t = \mathcal{S}_{\text{robust}}\left( \mathbf{f}_t, \text{Conf}_t, \Delta A_t \right),
\end{equation}
where $\text{Conf}_t$ measures prediction confidence and $\Delta A_t$ quantifies temporal mask stability. Features are registered as discriminative prototypes only when both ambiguity detection and quality assessment succeed:
\begin{equation}
	\mathcal{F}_{\text{disc}} \leftarrow \text{Update}(\mathcal{F}_{\text{disc}}, \mathbf{f}_t) \quad \text{s.t.} \quad \text{Ambiguous}(t) \land (s_t > \tau_{\text{reg}})
\end{equation}

At inference, the model queries all three dimensions simultaneously. Cross-modal semantic alignment anchors global identity for long-sequence consistency; short-term spatio-temporal evolution captures local appearance dynamics for adaptive tracking; discriminative refinement activates stored prototypes when encountering strong distractors. This synergistic design ensures memory reliability from three orthogonal dimensions---semantic stability, temporal adaptability, and distractor discriminability---significantly improving robustness and accuracy for long-sequence segmentation in complex remote sensing scenarios.

\section{Experiments}
\label{section:EXPERIMENTS}

\subsection{Dataset and Evaluation Metrics}

To comprehensively evaluate the proposed framework under the strict causality constraints inherent in remote sensing video analysis, all experiments are conducted on the RS-RVOS Bench dataset. As described in Section~\ref{subsection:dataset_statistics}, RS-RVOS Bench consists of 111 high-resolution satellite video sequences, covering approximately 25,000 frames and 213,000 temporal referring annotations, each associated with precise instance-level segmentation masks.

We strictly follow the official dataset split protocol, where 82 video sequences are used for training and the remaining 29 sequences are reserved for testing. The test set is carefully curated to include a diverse range of challenging scenarios, such as weak target saliency, dynamic occlusions, and complex background interference, enabling a rigorous evaluation of model robustness under realistic remote sensing conditions. All compared methods are trained exclusively on the training set, and quantitative results are reported on the test set without using any test-time augmentation or post-processing, ensuring fair and reproducible comparisons.

For evaluation, we adopt the standard metrics widely used in video object segmentation, assessing segmentation quality from complementary region-based and boundary-based perspectives~\cite{perazzi2016davis,ponttuset_davis17}. Specifically, region similarity $\mathcal{J}$, contour accuracy $\mathcal{F}$, and their mean $\mathcal{J}\&\mathcal{F}$ are employed as the primary quantitative indicators~\cite{perazzi2016davis}.

The region similarity metric $\mathcal{J}$ measures the intersection-over-Union (IoU) between the predicted mask $R_t$ and the ground-truth mask $G_t$~\cite{perazzi2016davis}. For a video sequence, $\mathcal{J}$ is computed as the average IoU over all frames containing the target object, defined as:
\begin{equation}
	\mathcal{J} = \frac{1}{|\mathcal{V}|} \sum_{t \in \mathcal{V}} \frac{|R_t \cap G_t|}{|R_t \cup G_t|},
\end{equation}
where $\mathcal{V}$ denotes the set of frames in which the target is present.

Given that targets in remote sensing imagery typically occupy only a small fraction of the frame and often exhibit complex boundaries, region-based metrics alone may be insufficient to capture boundary alignment quality. Therefore, we additionally employ the contour accuracy metric $\mathcal{F}$, which evaluates the agreement between predicted and ground-truth object boundaries. It is defined as the F-measure of contour precision $P_c$ and recall $R_c$:
\begin{equation}
	\mathcal{F} = \frac{2 \cdot P_c \cdot R_c}{P_c + R_c}
\end{equation}

During computation, a mild morphological dilation is applied to the contour maps to tolerate minor pixel-level misalignments, allowing the metric to focus on structural fidelity rather than exact pixel correspondence.

To provide a balanced assessment that jointly reflects region consistency and boundary accuracy, we further report the mean of $\mathcal{J}$ and $\mathcal{F}$ as the final evaluation score:
\begin{equation}
	\mathcal{J}\&\mathcal{F} = \frac{\mathcal{J} + \mathcal{F}}{2}
\end{equation}

This combined metric serves as the primary criterion for performance comparison and ranking in our experiments. 
All metrics are reported in the range $[0,1]$. Following the DAVIS protocol~\cite{perazzi2016davis,ponttuset_davis17}, we additionally report recall, defined as the fraction of frames whose per-frame score exceeds a fixed threshold $\tau_{\text{rec}}=0.5$:
\begin{equation}
	\begin{aligned}
		\text{Recall}_{\mathcal{J}} &= \frac{1}{|\mathcal{V}|}\sum_{t\in\mathcal{V}}\mathbb{I}\big[\mathcal{J}_t > \tau_{\text{rec}}\big], \\
		\text{Recall}_{\mathcal{F}} &= \frac{1}{|\mathcal{V}|}\sum_{t\in\mathcal{V}}\mathbb{I}\big[\mathcal{F}_t > \tau_{\text{rec}}\big]
	\end{aligned}
\end{equation}

\subsection{Experimental Setup}

The proposed MQC-SAM framework is built upon SAM2 (Hiera-L) as the video segmentation backbone under the online causal setting. A frozen BERT-base serves as the text encoder~\cite{bert_naacl19} to obtain language embeddings. CroBIM bridges language and SAM2 through cross-modal conditioning, enabling SAM2 to incorporate text-aware cues when propagating and refining masks across frames.

The model is trained on the RS-RVOS Bench training set using the AdamW optimizer~\cite{adamw_iclr19}, with a weight decay of 0.01 and an initial learning rate of $5 \times 10^{-5}$ following a polynomial decay schedule. Training is conducted for 40 epochs with a batch size of 32 on a computing cluster equipped with eight NVIDIA A800 GPUs. All components are trained in a unified manner on the RS-RVOS Bench training set unless otherwise stated.
To enable adaptive fusion of decoupled memory representations, the learnable attention fusion weights $\omega_1, \omega_2, \omega_3$ are all initialized to 1.0, while the relative positional bias matrix $\mathbf{P}{\text{rel}}$ is initialized to zero. Both are jointly optimized during training. In addition, the total capacity of the memory bank is constrained to $N{\max}=7$ to prevent excessive GPU memory consumption. To ensure input consistency, all video frames are uniformly resized to $480 \times 480$ pixels during both training and inference, and no data augmentation or post-processing techniques are applied.

\subsection{Comparison with State-of-the-art Methods}
\label{sec:cmp_sota}

We compare MQC-SAM with representative state-of-the-art RVOS methods from different paradigms, including transformer-based referring frameworks such as ReferFormer, online query-propagation baselines such as OnlineRefer, and large-model reasoning segmentation assistants such as VISA ~\cite{visa_eccv24}. We also include recent strong RVOS baselines (e.g., ReferDINO~\cite{referdino_iccv25} and SAMWISE~\cite{samwise_cvpr25}) as reported in Table~\ref{tab:sota_rvos}. For evaluation, we follow the classic DAVIS protocol where $\mathcal{J}$ measures region similarity (IoU) and $\mathcal{F}$ measures contour accuracy (boundary F-measure)~\cite{perazzi2016davis,ponttuset_davis17}. In Table~\ref{tab:sota_rvos}, we report J\&F-Mean, J-Mean, J-Recall, F-Mean, and F-Recall; The optimal and suboptimal performances are distinctly highlighted in \textcolor{red}{\textbf{red}} and \textcolor{blue}{\textbf{blue}}, respectively.

\begin{figure*}[t]
	\centering
	\includegraphics[width=\textwidth]{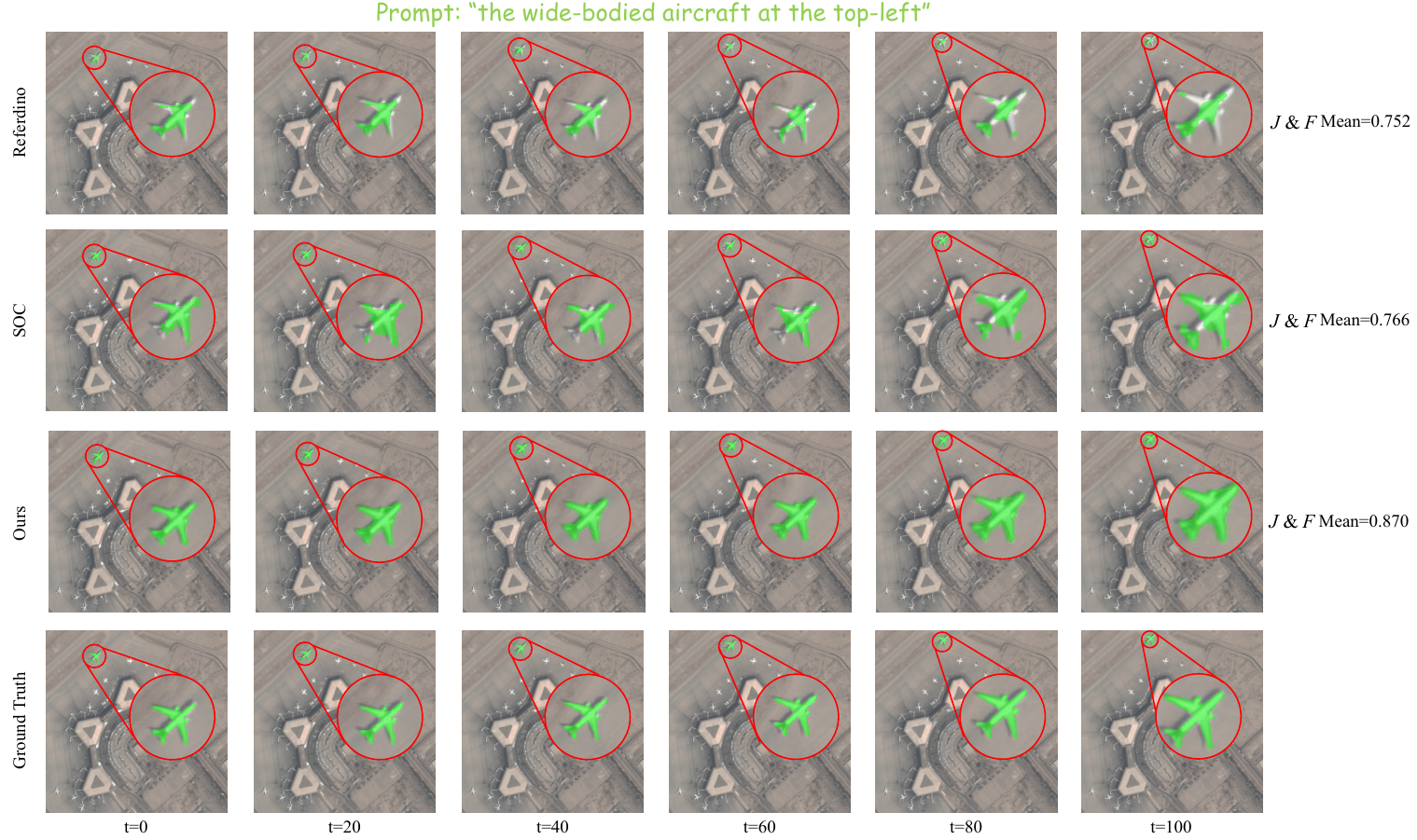}
	\caption{Qualitative comparison on a long-range remote sensing sequence with the prompt ``\emph{the wide-bodied aircraft at the top-left}''. Rows denote different methods, and columns are sampled frames over time. Green overlays indicate masks, and red annotations highlight the target region.
	}
	\label{fig:experiment1}
	
	\vspace{6pt} 
	
	\includegraphics[width=\textwidth]{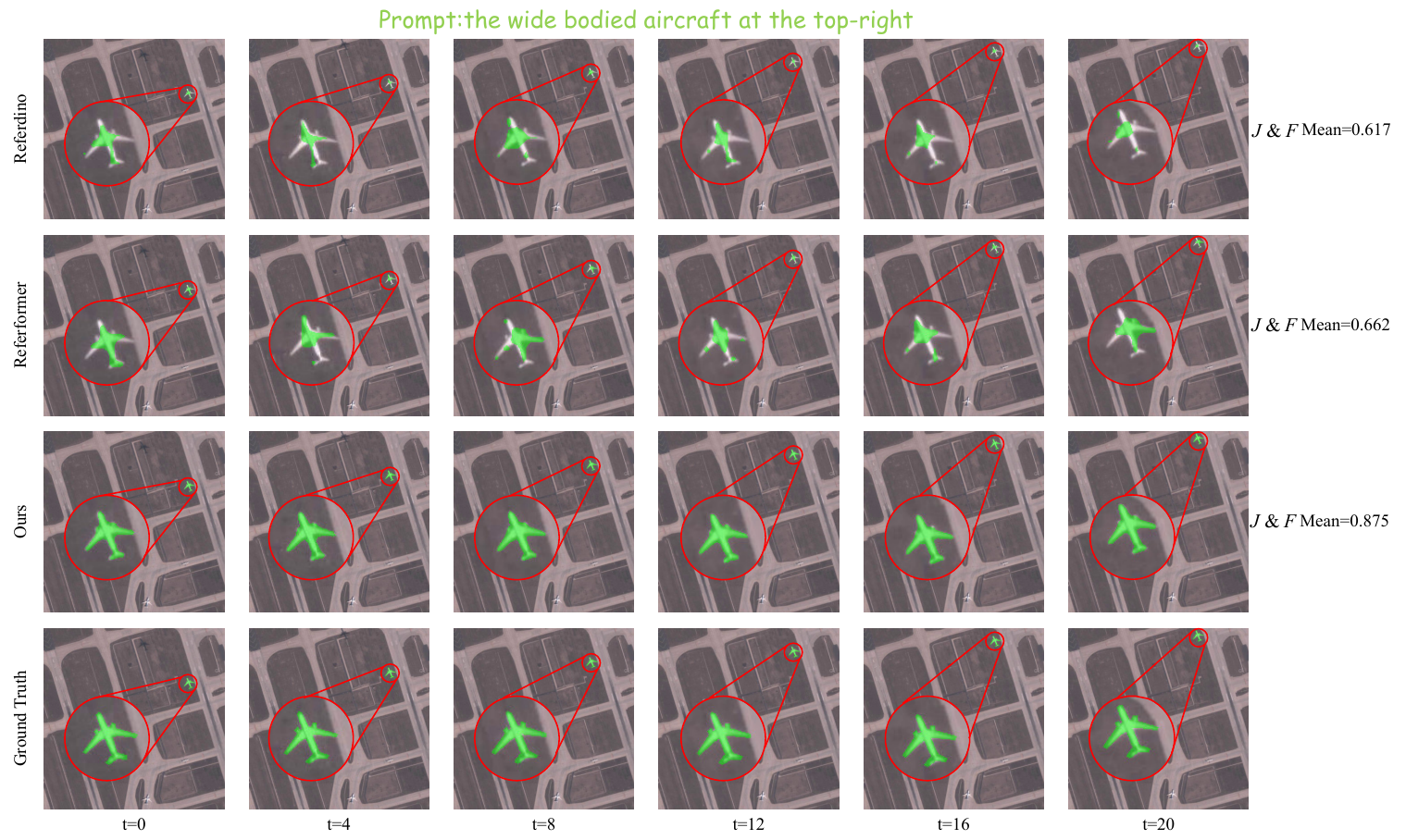}
	\caption{Qualitative comparison with the prompt ``\emph{the wide-bodied aircraft at the top-right}'' (same visualization setting as Figs.~\ref{fig:experiment1}).}
	\label{fig:experiment2}
\end{figure*}

\begin{figure*}[t]
	\centering
	
	\includegraphics[width=\textwidth]{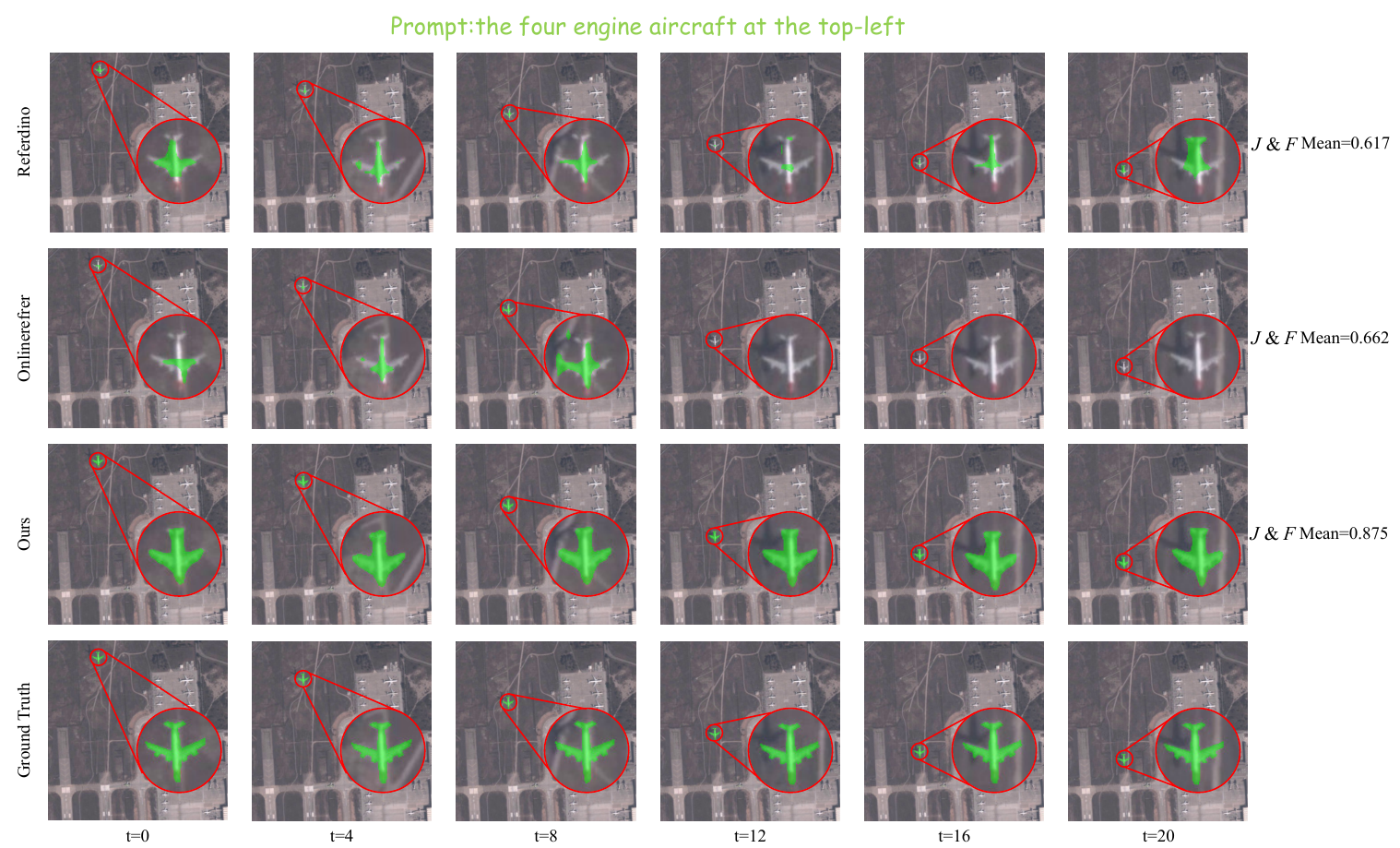}
	\caption{Qualitative comparison with the prompt ``\emph{the four engine aircraft at the top-left}'' (same visualization setting as Figs.~\ref{fig:experiment1}).}
	\label{fig:experiment3}
	
	\vspace{6pt}
	
	\includegraphics[width=\textwidth]{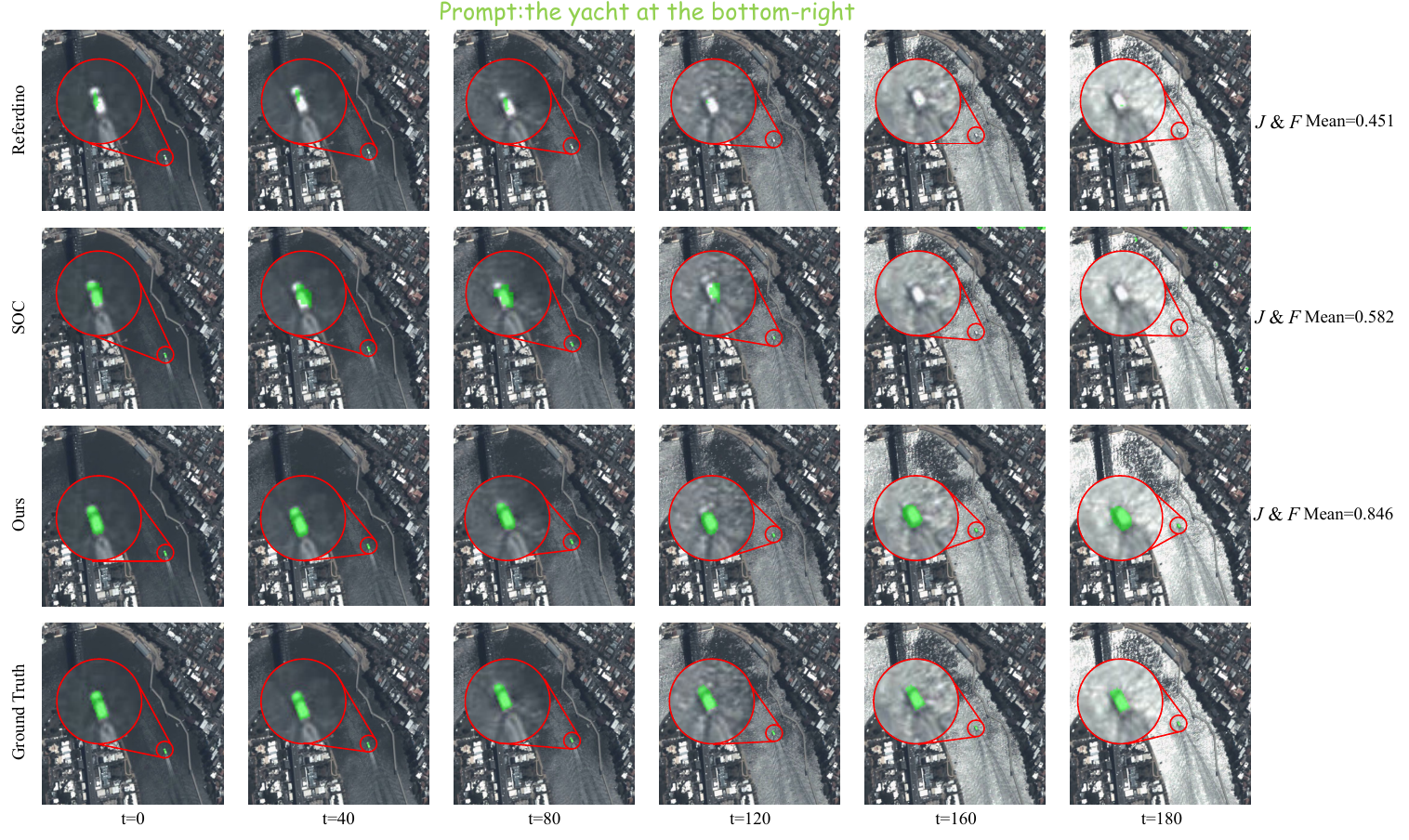}
	\caption{Qualitative comparison on a tiny object with the prompt ``\emph{the yacht at the bottom-right}'' (same visualization setting as Figs.~\ref{fig:experiment1}).}
	\label{fig:experiment4}
\end{figure*}

As shown in Table~\ref{tab:sota_rvos}, MQC-SAM achieves the best results across all metrics with a clear margin over prior methods. The improvement is particularly pronounced on J\&F-Mean and J-Recall, indicating that our method not only increases average region overlap but also substantially reduces catastrophic failures (e.g., target missing or severe drift) during temporal propagation. Meanwhile, the consistently higher F-Mean and F-Recall demonstrate that MQC-SAM better preserves accurate object boundaries, which is critical for tiny targets under weak saliency and cluttered backgrounds in remote sensing videos.
Figs.~\ref{fig:experiment1}--\ref{fig:experiment4} provide qualitative evidence that is consistent with the quantitative results Table~\ref{tab:sota_rvos}, clearly demonstrating the advantages of MQC-SAM in remote sensing RVOS scenarios. In Fig.~\ref{fig:experiment1}, we show a long-range sequence (up to \(t=100\)) with complex airport background. As time progresses, several competing methods gradually suffer from structural fragmentation, partial missing regions, and drift caused by confusion with background textures. In contrast, MQC-SAM consistently preserves target identity and produces more structurally complete and temporally coherent masks throughout the sequence.
Figs.~\ref{fig:experiment2} and \ref{fig:experiment3} further highlight the differences among methods on aircraft targets dominated by weak saliency. Some baselines tend to under-segment slender parts, resulting in thin, broken, or incomplete predictions that retain only a portion of the fuselage. When the target--background separability decreases in later frames, online propagation methods may also exhibit noticeable drift or even prediction collapse.

\clearpage

By comparison, MQC-SAM remains stably aligned with the referred target across different temporal stages, maintaining more complete region coverage and more consistent structural shapes. Fig.~\ref{fig:experiment4} presents an even more challenging tiny yacht under strong water-surface reflections and dynamic background interference. Competing methods are easily distracted by specular highlights and dynamic texture patterns, leading to fragmented, jittery, or unstable predictions. MQC-SAM, however, still produces compact and temporally consistent masks over a long horizon \(t \in [0, 180]\). 
Overall, these visual comparisons validate the effectiveness of our temporal motion-consistency initial memory calibration and the decoupled attention memory integration strategy, thereby improving robustness and accuracy for remote sensing RVOS.

\begin{table*}[t]
	\centering
	\caption{Performance Comparison on RS-RVOS Bench Dataset. Optimal values are highlighted in \textcolor{red}{\textbf{bold red}} and sub-optimal values in \textcolor{blue}{\textbf{bold blue}}.}
	\label{tab:sota_rvos}
	\renewcommand{\arraystretch}{1.12}
	\setlength{\tabcolsep}{6pt}
	
	\begin{threeparttable}
		\resizebox{\textwidth}{!}{%
			\begin{tabular}{l| l l l| c c c c c}
				\hline\hline
				Method & Venue & Text Encoder & Visual Encoder
				& J\&F$\uparrow$ & J$\uparrow$ & J-Recall$\uparrow$ & F$\uparrow$ & F-Recall$\uparrow$ \\
				\hline
				ReferFormer~\cite{wu2022referformer} & CVPR 2022 & RoBERTa & Video-Swin-B
				& 0.526 & 0.296 & 0.196 & 0.756 & 0.811 \\
				OnlineRefer~\cite{wu2023onlinerefer} & ICCV 2023 & RoBERTa & Swin-B
				& 0.324 & 0.149 & 0.083 & 0.500 & 0.551 \\
				VISA~\cite{visa_eccv24} & ECCV 2024 & Chat-UniVi-7B & ViT-H
				& 0.215 & 0.118 & 0.014 & 0.312 & 0.241 \\
				SAMWISE~\cite{samwise_cvpr25} & CVPR 2025 & RoBERT & Hiera-L
				& 0.481 & 0.317 & \textcolor{blue}{\textbf{0.306}} & 0.645 & 0.682 \\
				SOC~\cite{luo2023soc} & NeurIPS 2023 & RoBERTa & Video-Swin-B
				& 0.539 & \textcolor{blue}{\textbf{0.348}} & 0.267 & 0.730 & 0.761 \\
				ReferDINO~\cite{referdino_iccv25} & ICCV 2025 & BERT & Swin-T
				& 0.519 & 0.285 & 0.215 & 0.753 & 0.805 \\
				ReferDINO~\cite{referdino_iccv25} & ICCV 2025 & BERT & Swin-B
				& \textcolor{blue}{\textbf{0.562}} & 0.303 & 0.210 & \textcolor{blue}{\textbf{0.821}} & \textcolor{blue}{\textbf{0.880}} \\
				\hline
				\textbf{Ours} & \multicolumn{1}{c}{-} & BERT & Hiera-L
				& \textcolor{red}{\textbf{0.712}} & \textcolor{red}{\textbf{0.506}} & \textcolor{red}{\textbf{0.541}}
				& \textcolor{red}{\textbf{0.918}} & \textcolor{red}{\textbf{0.952}} \\
				\hline\hline
			\end{tabular}%
		}
		
		\begin{tablenotes}[flushleft]
			\footnotesize
			\item \textit{Note:} Arrows indicate the desired direction of improvement ($\downarrow$ lower is better, $\uparrow$ higher is better).
		\end{tablenotes}
	\end{threeparttable}
\end{table*}

\begin{table*}[t]
	\centering
	\caption{ABLATION STUDY OF KEY COMPONENTS IN MQC-SAM ON THE RS-RVOS BENCH DATASET.}
	\label{tab:ablation_rsrvos}
	
	\setlength{\tabcolsep}{5.5pt}
	\renewcommand{\arraystretch}{1.12}
	
	\begin{tabularx}{\textwidth}{c>{\raggedright\arraybackslash}Xccccc}
		\toprule \toprule
		\multirow{2}{*}{ID} & \multirow{2}{*}{Method}
		& \multirow{2}{*}{J\&F-Mean$\uparrow$}
		& \multicolumn{2}{c}{$\mathcal{J}$ (Region)$\uparrow$}
		& \multicolumn{2}{c}{$\mathcal{F}$ (Contour)$\uparrow$} \\
		\cmidrule(lr){4-5} \cmidrule(lr){6-7}
		& & & J-Mean & J-Recall & F-Mean & F-Recall \\
		\midrule
		
		1 & Baseline
		& 0.655 & 0.450 & 0.441 & 0.859 & 0.893 \\
		
		2 & Baseline + \textbf{TMCC}
		& 0.699 \textbf{\textcolor{green!50!black}{\scriptsize{($+0.044$)}}}
		& 0.480 \textbf{\textcolor{green!50!black}{\scriptsize{($+0.030$)}}}
		& 0.482 \textbf{\textcolor{green!60!black}{\scriptsize{($+0.041$)}}}
		& 0.917 \textbf{\textcolor{green!50!black}{\scriptsize{($+0.058$)}}}
		& 0.951 \textbf{\textcolor{green!50!black}{\scriptsize{($+0.058$)}}} \\
				
		3 & Baseline + \textbf{DAMI}
		& 0.678 \textbf{\textcolor{green!50!black}{\scriptsize{($+0.023$)}}}
		& 0.457 \textbf{\textcolor{green!60!black}{\scriptsize{($+0.007$)}}}
		& 0.446 \textbf{\textcolor{green!60!black}{\scriptsize{($+0.005$)}}}
		& 0.899 \textbf{\textcolor{green!50!black}{\scriptsize{($+0.040$)}}}
		& 0.955 \textbf{\textcolor{green!50!black}{\scriptsize{($+0.062$)}}} \\
		
		4 & \textbf{Full}
		& \textbf{0.712} \textbf{\textcolor{green!50!black}{\scriptsize{($+0.057$)}}}
		& \textbf{0.506} \textbf{\textcolor{green!50!black}{\scriptsize{($+0.056$)}}}
		& \textbf{0.541} \textbf{\textcolor{green!50!black}{\scriptsize{($+0.100$)}}}
		& \textbf{0.918} \textbf{\textcolor{green!50!black}{\scriptsize{($+0.059$)}}}
		& 0.952 \textbf{\textcolor{green!50!black}{\scriptsize{($+0.059$)}}} \\
		
		\bottomrule \bottomrule
	\end{tabularx}
\end{table*}

\begin{figure*}[t]
	\centering
	\includegraphics[width=\textwidth]{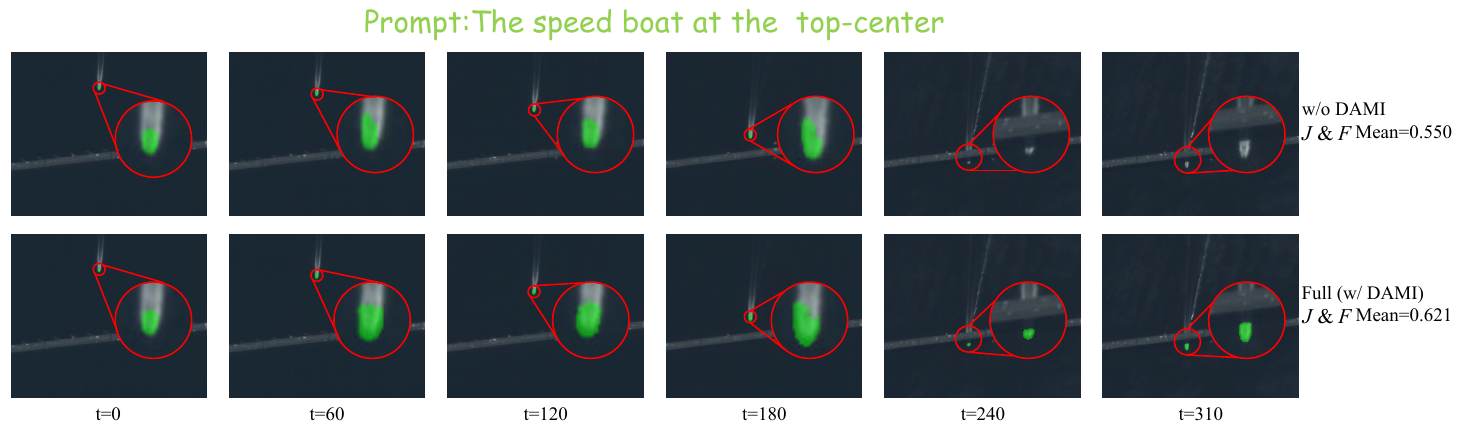}
	\caption{Qualitative ablation on removing DAMI. Top: Full model (Baseline+TMCC+DAMI); Bottom: w/o DAMI (Baseline+TMCC). DAMI improves temporal stability for tiny targets under weak saliency, reducing mask shrinkage and intermittent missing in later frames.}
	\label{fig:ablation1}
\end{figure*}

\subsection{Ablation Study}
\label{sec:ablation}

We conduct ablation studies on the RS-RVOS Bench dataset to quantify the contribution of our two key components: TMCC and DAMI. Following the classic DAVIS protocol, we report J\&F-Mean, J-Mean, J-Recall, F-Mean, and F-Recall, where higher values indicate better performance. The quantitative results are summarized in Table~\ref{tab:ablation_rsrvos}.

Starting from the baseline (CroBIM+SAM2, ID~1), incorporating TMCC (ID~2) brings a substantial improvement, raising J\&F-Mean from 0.655 to 0.699. In particular, both region-based metrics improve consistently (J-Mean: 0.450$\rightarrow$0.480; J-Recall: 0.441$\rightarrow$0.482), and contour quality increases markedly (F-Mean: 0.859$\rightarrow$0.917; F-Recall: 0.893$\rightarrow$0.951). These gains indicate that motion-consistency calibration effectively refines the initial mask and constructs more reliable initial memory, which reduces background contamination and mitigates error propagation in subsequent frames.

Adding DAMI alone (ID~3) also improves performance over the baseline, increasing J\&F-Mean to 0.678 and boosting F-Recall to 0.955. This suggests that the decoupled tri-attention memory integration enhances retrieval robustness under distractors and weak saliency by leveraging complementary cues from semantic anchoring, short-term evolution, and discriminative refinement. When combining TMC and DAMI (Full model, ID~4), we obtain the best overall results across all metrics, achieving 0.712 in J\&F-Mean and consistently higher J/F scores, which validates the complementarity between a stronger initialization (TMCC) and a more reliable long-term memory integration mechanism (DAMI).

In addition to quantitative results, Fig.~\ref{fig:ablation1} provides a qualitative ablation by removing DAMI. In this sequence, the speed boat travels under a bridge, leading to a clear occlusion period where the target becomes barely visible, causing a temporal visual interruption. Without DAMI, the model is more likely to lose target identity after the occlusion and fails to reliably re-lock onto the boat when it reappears, resulting in unstable or missing predictions in later frames. In contrast, the full model remains robust to such interruption: it preserves a stable target memory during occlusion and quickly recovers consistent segmentation once the target becomes visible again, producing compact and temporally coherent masks throughout the sequence. This observation highlights that DAMI substantially improves long-term identity maintenance under occlusion and intermittent visibility, which aligns with the recall gains reported in Table~\ref{tab:ablation_rsrvos}.

\section{Conclusion}
\label{sec:conclusion}

This paper investigates RVOS in remote sensing scenarios. Unlike natural-scene RVOS, remote sensing videos are typically characterized by extremely weak target saliency, complex background structures, and intermittent visibility caused by occlusions from clouds or large man-made structures. These factors significantly amplify drift and error accumulation during temporal propagation, leading to unstable target identity and degraded segmentation quality. To facilitate research and evaluation in this direction, we build the RS-RVOS Bench dataset, providing a unified benchmark and quantitative evaluation platform for remote sensing RVOS.

To address the above challenges, we propose MQC-SAM, a two-stage memory management framework tailored for remote sensing RVOS. First, we introduce TMCC to validate and refine the initial mask using temporally accumulated motion evidence, producing a cleaner and more reliable initial memory. Second, we develop DAMI, which decomposes memory integration into three orthogonal attention dimensions to jointly achieve stable identity anchoring, short-term appearance adaptation, and discriminative refinement against confusing distractors. Together, these designs enforce reliable memory registration and effectively suppress progressive drift in long-term temporal reasoning.

Extensive experiments on RS-RVOS Bench demonstrate that MQC-SAM consistently improves over state-of-the-art RVOS methods under the classic $\mathcal{J}/\mathcal{F}$ evaluation protocol. Qualitative results further show that MQC-SAM maintains structurally complete masks and stable target correspondence over long sequences, and remains robust under weak-saliency small targets and complex background interference. Ablation studies validate the effectiveness and complementarity of TMCC and DAMI; in particular, DAMI strengthens identity maintenance under occlusion-induced temporal interruptions, enabling reliable re-locking and recovery after the target reappears.

In future work, we will explore more efficient memory compression and adaptive update strategies for ultra-long remote sensing videos, and extend the proposed framework to broader remote sensing spatiotemporal understanding tasks with more complex targets and scene dynamics.

\ifCLASSOPTIONcaptionsoff
\newpage
\fi
\bibliography{refs}
\vspace{-10 mm} 
\begin{IEEEbiography}[{\includegraphics[width=1in,height=1.25in,clip,keepaspectratio]{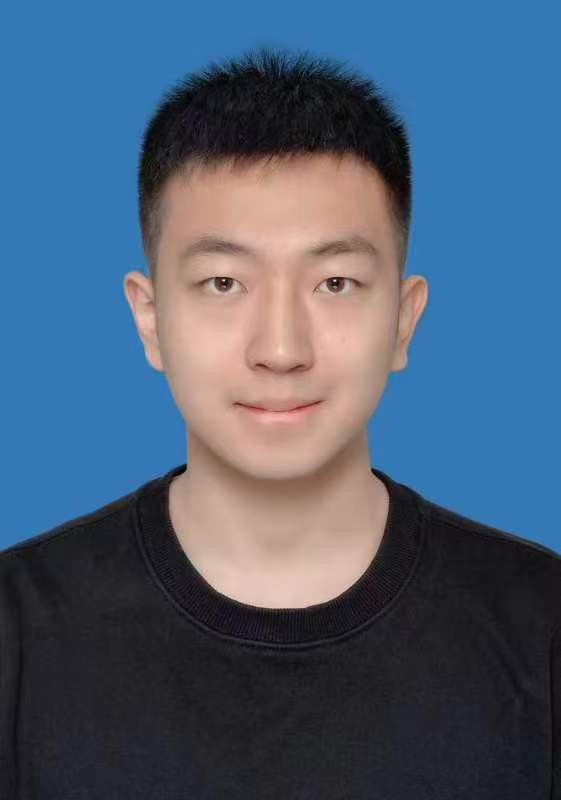}}]{Haochen Jiang}  received his bachelor's degree in Electronic Information Science and Technology from Jilin University, China, and is currently pursuing a Ph.D. degree in Information and Communication Engineering. His research interests include multimodal learning with remote sensing imagery and text, downstream tasks for remote sensing image understanding, and the development of foundation models in remote sensing.
\end{IEEEbiography}
\vspace{-10 mm} 
\begin{IEEEbiography}[{\includegraphics[width=1in,height=1.25in,clip,keepaspectratio]{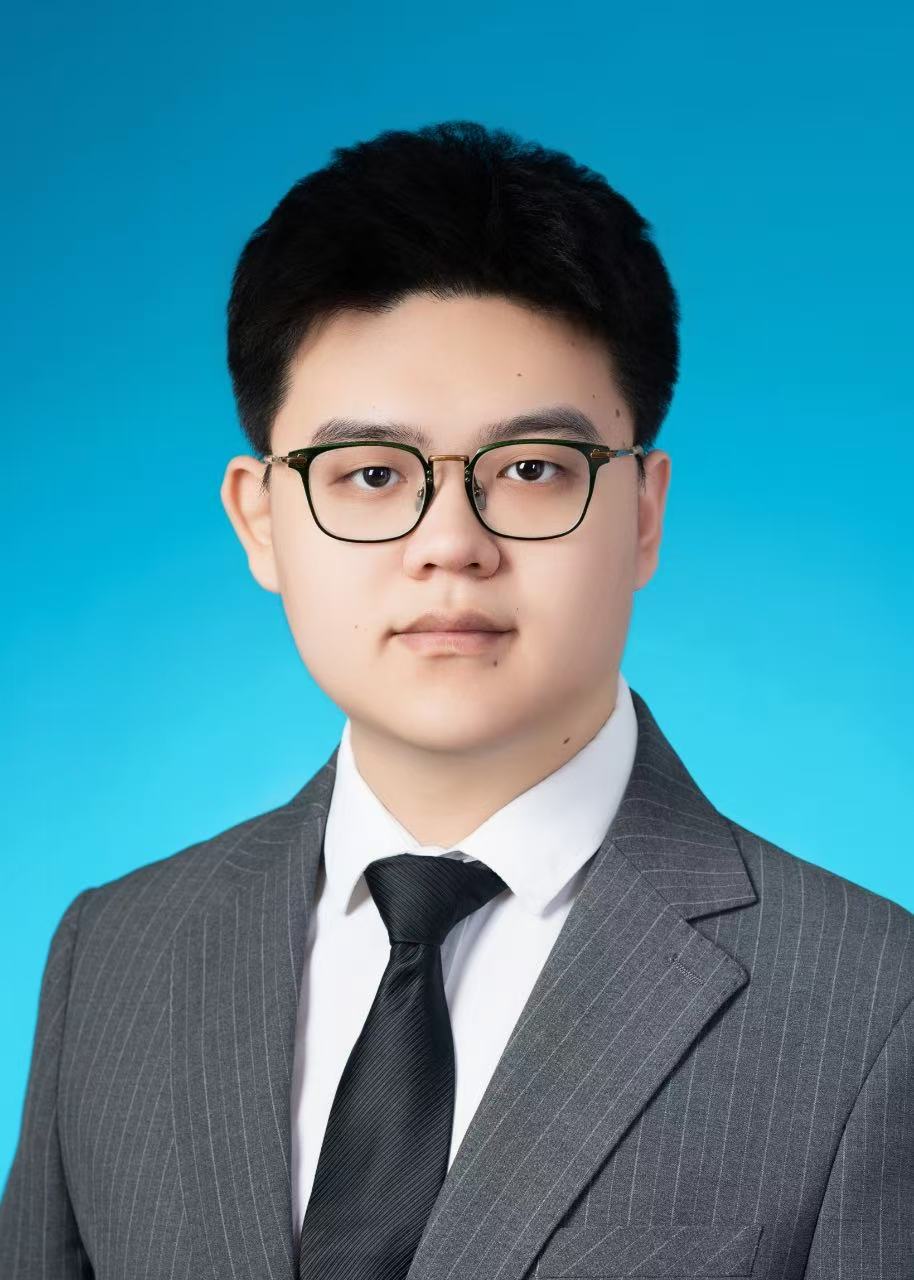}}]{Yuzhe Sun} received his bachelor's degree in Remote Sensing Science and Technology from Harbin Institute of Technology, China, and is currently pursuing the Ph.D. degree in Information and Communication Engineering. His research interests include the development of cross-modal remote sensing models for image and text, downstream tasks of remote sensing images, and the construction of foundational models for remote sensing.
\end{IEEEbiography}
\vspace{-10 mm} 
\begin{IEEEbiography}[{\includegraphics[width=1in,height=1.25in,clip,keepaspectratio]{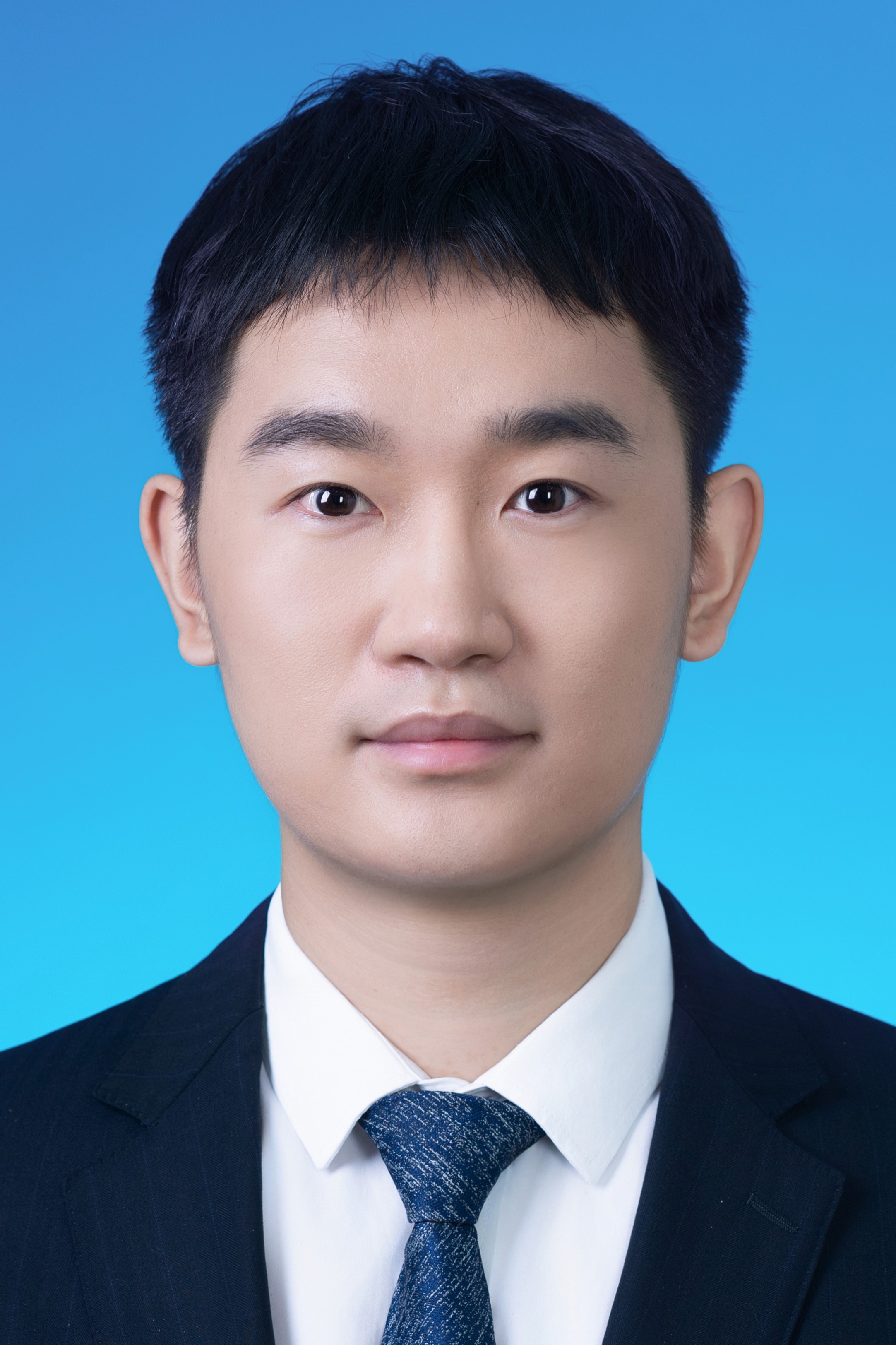}}]{Zhe Dong} received the M.S. degree in software engineering from Harbin Engineering University, Harbin, China, in 2022. He is currently pursuing the Ph.D. degree with the School of Electronics and Information Engineering, Harbin Institute of Technology, Harbin. His research interests are related to the semantic segmentation and self-supervised learning of remote sensing images. 
\end{IEEEbiography}
\vspace{-10 mm} 
\begin{IEEEbiography}[{\includegraphics[width=1in,height=1.25in,clip,keepaspectratio]{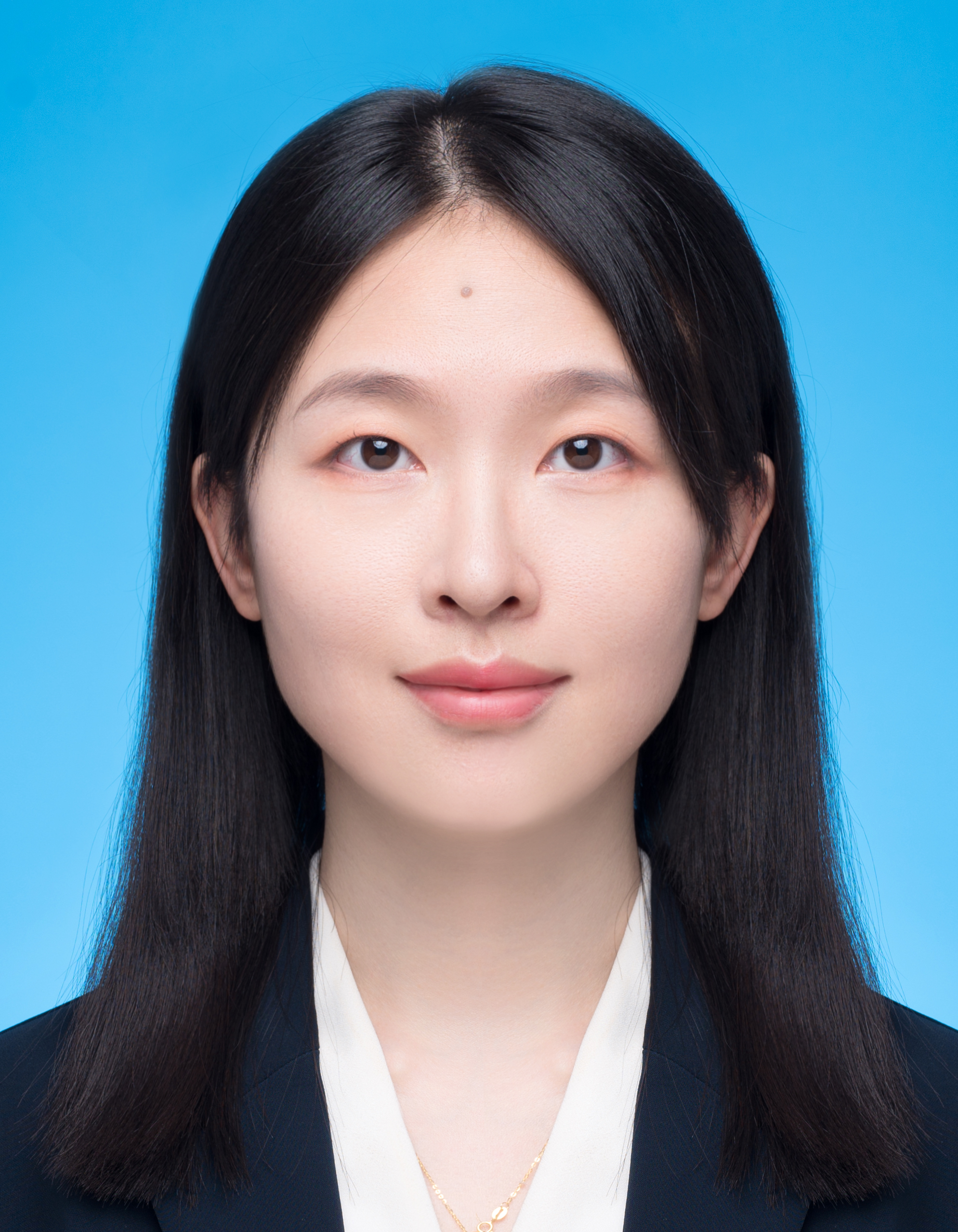}}]{Tianzhu Liu} (Member, IEEE) received the Ph.D. degree in information and communication engineering from the Harbin Institute of Technology (HIT), Harbin, China, in 2019. She was a Lecturer with the School of Electronics and Information Engineering, HIT, where she is currently an Associate Professor and a Post-Doctoral Fellow. Her research interests include image processing in hyperspectral remote sensing, especially multimodal hyperspectral image classification. Dr. Liu won the Third Prize in Student Paper Contest at the International Geoscience and Remote Sensing Symposium (IGARSS) in 2019. She serves as an Associate Editor for the \textit{IEEE Journal of Selected Topics in Applied Earth Observations and Remote Sensing} and a Reviewer for several international journals, such as the \textit{IEEE Transactions on Geoscience and Remote Sensing} and \textit{Neurocomputing}.
\end{IEEEbiography}
\vspace{-10 mm} 
\begin{IEEEbiography}[{\includegraphics[width=1in,height=1.25in,clip,keepaspectratio]{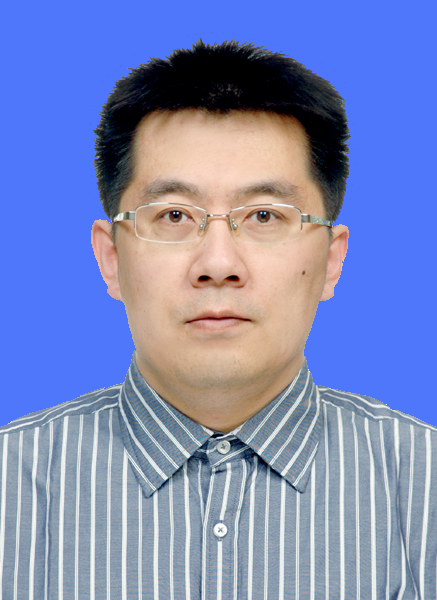}}]{Yanfeng Gu} (M’06-SM’16) received the Ph.D. degree in information and communication engineering from Harbin Institute of Technology, Harbin, China, in 2005. He joined as a Lecture with the School of Electronics and Information Engineering, Harbin Institute of Technology (HIT). He was appointed as Associate Professor at the same institute in 2006; meanwhile, he was enrolled in first Outstanding Young Teacher Training Program of HIT. From 2011 to 2012, he was a Visiting Scholar with the Department of Electrical Engineering and Computer Science, University of California, Berkeley, CA, USA. He is currently a Professor with the Department of Information Engineering, HIT, Harbin, China. He has published more than 100 peer-reviewed papers, four book chapters, and he is the inventor or coinventor of 20 patents. His research interests include space intelligent remote sensing and information processing, multimodal hyperspectral remote sensing, spaceborne time-series image processing.
\end{IEEEbiography}

\end{document}